\documentclass[runningheads]{llncs}
\usepackage{graphicx}
\usepackage{comment}
\usepackage[scaled=.90]{helvet}
\usepackage{courier}
\usepackage{amsmath,amssymb} %
\usepackage{color}
\usepackage{epsfig}
\usepackage{graphicx}
\usepackage{amsmath}
\usepackage{bm}
\usepackage{subcaption}
\usepackage{comment}
\usepackage{amssymb}
\usepackage{indentfirst}
\usepackage[font=small,labelfont=bf]{caption}
\usepackage{xfrac}

\usepackage{mmstyle}
\usepackage{pifont}
\usepackage{multirow}

\newcommand{\cmark}{\ding{51}}
\usepackage{wrapfig}
\usepackage{url}

\newcommand{\vs}{\textit{vs.}}

\usepackage[pagebackref=true,breaklinks=true,letterpaper=true,colorlinks,bookmarks=false]{hyperref}

\begin{document}

\pagestyle{headings}
\mainmatter

\title{Omni-sourced Webly-supervised Learning for Video Recognition}

\author{Haodong Duan\inst{1} \and
Yue Zhao\inst{1} \and Yuanjun Xiong\inst{2} \and Wentao Liu\inst{3} \and
Dahua Lin\inst{1}}
\institute{The Chinese University of Hong Kong \and Amazon AI \and Sensetime Research}

\maketitle

\begin{abstract}
We introduce OmniSource, a novel framework for leveraging web data to train video recognition models. 
OmniSource overcomes the barriers between data formats, such as images, short videos, and long untrimmed videos for webly-supervised learning.
First, data samples with multiple formats, curated by task-specific data collection and automatically filtered by a teacher model, are transformed into a unified form. 
Then a joint-training strategy is proposed to deal with the domain gaps between multiple data sources and formats in webly-supervised learning. 
Several good practices, including data balancing, resampling, and cross-dataset mixup are adopted in joint training.
Experiments show that by utilizing data from multiple sources and formats, OmniSource is more data-efficient in training.
With only 3.5M images and 800K minutes videos crawled from the internet without human labeling (less than $2\%$ of prior works), our models learned with OmniSource improve Top-1 accuracy of 2D- and 3D-ConvNet baseline models by 3.0\% and 3.9\%, respectively, on the Kinetics-400 benchmark.
With OmniSource, we establish new records with different pretraining strategies for video recognition.
Our best models achieve \textbf{80.4\%}, \textbf{80.5\%}, and \textbf{83.6\%} Top-1 accuracies on the Kinetics-400 benchmark respectively for training-from-scratch, ImageNet pre-training and IG-65M pre-training.
\end{abstract}

\section{Introduction}
Following the great success of representation learning in image recognition~\cite{krizhevsky2012imagenet,simonyan2014very,he2016deep,huang2017densely}, recent years have witnessed great progress in video classification thanks to the development of stronger models~\cite{simonyan2014two,wang2018temporal,carreira2017quo,tran2019video} as well as the collection of larger-scale datasets~\cite{carreira2017quo,zhao2019hacs,monfort2019moments,miech2019howto100m}.
However, labelling large-scale image datasets~\cite{russakovsky2015imagenet,zhou2017places} is well known to be costly and time-consuming.
It is even more difficult to do so for trimmed video recognition.
The reason is that most online videos are \emph{untrimmed},~\ie containing numerous shots with multiple concepts, making it unavoidable to first go through the entire video and then manually cut it into informative video clips based on a specific query.
Such procedure requires far more efforts than image annotation where a simple glance and click is needed.
As a result, while the quantity of web videos grows exponentially over the past 3 years, the Kinetics dataset merely grows from 300K videos in 400 classes~\cite{kay2017kinetics} to 650K in 700 classes~\cite{carreira2019short}, partially limiting the scaling-up of video architectures~\cite{carreira2017quo}.

\noindent
\begin{figure}
	\begin{center}
		\includegraphics[width=.8\linewidth]{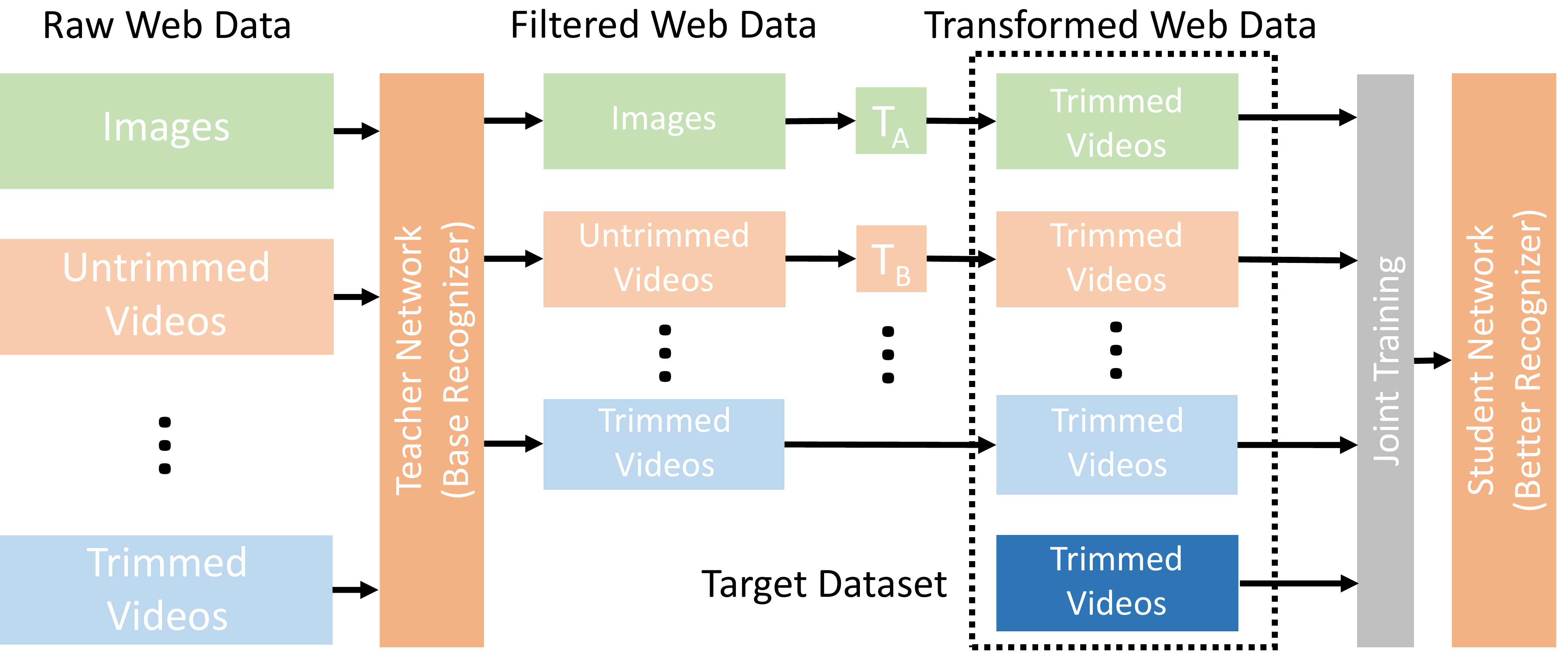}
	\end{center}
	\captionof{figure}{\textbf{OmniSource Framework.} We first train a teacher network on the target dataset. Then, we use the teacher network to filter collected web data of different formats, to reduce noise and improve data quality. Specific transformations are conducted on the filtered out data corresponding to their formats. The target dataset and auxiliary web datasets are used for joint training of the student network}
	\label{fig:tether}
\end{figure}

Instead of confining ourselves to the well-annotated trimmed videos, we move beyond by exploring the abundant visual data that are publicly available on the Internet in a more labor-saving way.
These visual data are in various formats, including images, short video clips, and long videos.
They capture the same visual world while exhibiting different advantages: \eg images may be of higher quality and focus on distinctive moments;
short videos may be edited by the user, therefore contain denser information;
long videos may depict an event in multiple views.
We transform the data in different formats into a unified form so that a single model can combine the best of both worlds.

Recent works~\cite{mahajan2018exploring,ghadiyaram2019large} explore the possibility of pre-training from massive unlabeled web images or videos only with hashtags.
However, they restrict the scope to the data of a single format.
Also, these methods usually require billions of images to obtain a pre-trained 2D CNN model that is resilient to noise, which poses great costs and restricts its practicability.
Besides, to take advantage of representation learned from large-scale images for videos, we have to take extra steps to transfer the 2D ConvNets to the 3D counterparts, either by inflating~\cite{carreira2017quo} or distillation~\cite{girdhar2019distinit}, and then perform fine-tuning on the target dataset, which is tedious and may be suboptimal.

In this work, we propose a simple and unified framework for video classification while utilizing multiple sources of web data in different formats simultaneously.
To enhance data efficiency, we propose task-specific data collection,~\ie obtaining topmost results using class labels as keywords on search engines, making the supervision most informative.
Our framework consists of three steps:
(1) We train one (or more) teacher network on the labeled dataset;
(2) For each source of data collected, we apply the corresponding teacher network to obtain pseudo-labels and filter out irrelevant samples with low confidence;
(3) We apply different transforms to convert each type of web data (\eg images) to the target input format (\eg video clips) and train the student network.

There are two main obstacles during joint training with the labeled dataset and unlabeled web datasets.
First, possible domain gaps occur. 	
For example, web images may focus more on objects and contain less motion blur than videos.
Second, teacher filtering may lead to unbalanced data distribution across different classes.
To mitigate the domain gap, we propose to balance the size of training batches between the labeled dataset and unlabeled web datasets and apply cross-dataset \emph{mixup}.
To cope with data imbalance, we try several resampling strategies.
All these techniques contribute to the success of our approach.

Compared to the previous methods, our method excels at the following aspects:
(1) It leverages a mixture of web data forms, including images, trimmed videos and untrimmed videos into one student network, aiming at an \emph{omni-sourced} fashion.
(2) It is \emph{data-efficient}.
Empirical results show that only 2M images, a significantly smaller amount compared to the total frame number of Kinetics (240K videos, $\sim$ 70M frames), are needed to produce notable improvements (about 1\%).
For trimmed videos, the required amount is around 0.5M.
In stark contrast, 65M videos are collected to obtain a noise-resilient pre-trained model in~\cite{ghadiyaram2019large,yalniz2019billion}.
It is also noteworthy that our framework can also benefit from the massively weakly-supervised pre-training from billions of images or videos.

To sum up, our contributions are as follows:

(1) We propose OmniSource, a simple and efficient framework for webly-supervised video classification, which can leverage web data in different formats.

(2) We propose good practices for problems during joint training with omni-sourced data, include source-target balancing, resampling and cross-dataset mixup.

(3) In experiments, our models trained by OmniSource achieve state-of-the-art performance on the Kinetics-400, for all pre-training strategies we tested.

\section{Related work}

\noindent\textbf{Webly-supervised learning}
Leveraging information from the Internet, termed \emph{webly-supervised learning}, has been extensively explored~\cite{lee2018cleannet,yang2018recognition,guo2018curriculumnet,gan2016you}.
Divvala~\etal in~\cite{divvala2014learning} proposes to automatically learn models from online resources for visual concept discovery and image annotation.
Chen~\etal reveals that images crawled from the Internet can yield superior results over the fully-supervised method~\cite{chen2015webly}.
For video classification, Ma~\etal proposes to use web images to boost action recognition models in~\cite{ma2017less} at the cost of manually filtering web action images.
To free from additional human labor, efforts have been made to learn video concept detectors~\cite{ye2015eventnet,liang2016learning} or to select relevant frames from videos~\cite{gan2016webly,sun2015temporal,yeung2017learning}.
These methods are based on frames thus fail to consider the rich temporal dynamics of videos.
Recent works~\cite{mahajan2018exploring,ghadiyaram2019large} show that webly-supervised learning can produce better pre-training models with very large scale noisy data ($\sim10^9$ images and $\sim10^7$ videos).
Being orthogonal to the pre-training stage, our framework works in a joint-training paradigm and is complementary to large-scale pre-training.

\noindent\textbf{Semi-supervised learning}
Our framework works under the semi-supervised setting where labeled and unlabeled(web) data co-exist.
Representative classical approaches include label propagation~\cite{zhu2002learning}, self-training~\cite{rosenberg2005semi}, co-training~\cite{blum1998combining}, and graph networks~\cite{kipf2016semi}.
Deep models make it possible to learn directly from unlabeled data via generative models~\cite{kingma2014semi}, self-supervised learning~\cite{zhai2019s4l}, or consensus of multiple experts~\cite{zhan2018consensus}.
However, most existing methods are validated only on small scale datasets.
One concurrent work~\cite{yalniz2019billion} proposes to first train a student network with unlabeled data with pseudo-labels and then fine-tune it on the labeled dataset.
Our framework, however, works on the two sources simultaneously, free from the pretrain-finetune paradigm and is more data-efficient.

\noindent\textbf{Distillation}
According to the setting of knowledge distillation~\cite{hinton2015distilling} and data distillation~\cite{radosavovic2018data}, given a set of manually labeled data, we can train a base model in the manner of supervised learning. 
The model is then applied to the unlabeled data or its transforms.
Most of the previous efforts~\cite{radosavovic2018data} are confined to the domain of images.
In~\cite{girdhar2019distinit}, Rohit~\etal proposes to distill spatial-temporal features from unlabeled videos with image-based teacher networks.
Our framework is capable of distilling knowledge from multiple sources and formats within a single network.

\noindent\textbf{Domain Adaptation} 
Since web data from multiple sources are taken as input, domain gaps inevitably exist.
Previous efforts~\cite{csurka2017comprehensive,tzeng2017adversarial,chen2019temporal} in domain adaptation focus on mitigating the data shift~\cite{quionero2009dataset} in terms of data distributions.
On the contrary, our framework focuses on adapting visual information in different formats (\eg still images, long videos) into the same format (\ie trimmed video clips).

\noindent\textbf{Video classification}
Video analysis has long been tackled using hand-crafted feature ~\cite{laptev2005space,wang2013action}.
Following the success of deep learning for images, video classification architectures have been dominated by two families of models,~\ie two-stream ~\cite{simonyan2014two,wang2018temporal} and 3D ConvNets~\cite{carreira2017quo,tran2018closer}.
The former uses 2D networks to extract image-level feature and performs temporal aggregation~\cite{wang2018temporal,zhou2018temporal,hussein2019timeception} on top while the latter learns spatial-temporal features directly from video clips~\cite{tran2018closer,feichtenhofer2019slowfast,tran2019video}.

\section{Method}

\subsection{Overview}
We propose a unified framework for omni-sourced webly-supervised video recognition, formulated in Sec.~\ref{subsec:formulation}.
The framework exploits web data of various forms (images, trimmed videos, untrimmed videos) from various sources (search engine, social media, video sharing platform) in an integrated way.
Since web data can be very noisy, we use a teacher network to filter out samples with low confidence scores and obtain pseudo labels for the remaining ones (Sec.~\ref{subsec:teacher_filter}).
We devise transformations for each form of data to make them applicable for the target task in Sec.~\ref{subsec:transform}.
In addition, we explore several techniques to improve the robustness of joint training with web data in Sec.~\ref{subsec:joint_train}.

\subsection{Framework formulation}
\label{subsec:formulation}
Given a target task (trimmed video recognition, e.g.) and its corresponding \emph{target} dataset $ \cD_\cT = \{ (\vx_i, \vy_i) \} $, we aim to harness information from unlabeled web resources $ \cU = \cU_1 \cup \cdots \cup \cU_n $, where $ \cU_i $ refers to unlabeled data in a specific source or format.
\textbf{First}, we construct the pseudo-labeled dataset $ \widehat{\cD}_i $ from $ \cU_i $.
Samples with low confidence are dropped using a teacher model $ \cM $ trained on $ \cD_\cT $,
and the remaining data are assigned with pseudo-labels $ \widehat\vy = \text{PseudoLabel}(\cM(\vx)) $.
\textbf{Second}, we devise appropriate transforms $ \cT_i (\vx): \widehat{\cD}_i \rightarrow \cD_{\cA, i} $ to process data in a specific format (\eg still images or long videos) into the data format (trimmed videos in our case) in the target task.
We denote the union of $ \cD_{\cA, i} $ to be the \emph{auxiliary} dataset $ \cD_\cA $.
\textbf{Finally}, a model $ \cM' $ (not necessarily the original $ \cM $), can be jointly trained on $ \cD_\cT $ and $ \cD_\cA $.
In each iteration, we sample two mini-batches of data $\cB_\cT$, $\cB_\cA$ from $\cD_\cT$, $\cD_\cA$ respectively.
The loss is a sum of cross entropy loss on both $\cB_\cT$ and $\cB_\cA$, indicated by Eq~\ref{eq:loss}.
\begin{equation}
\cL = \sum_{ \vx, \vy \in \cB_{\cT}} \cL(\cF(\vx; \cM'), \vy) +  \sum_{ \vx, \widehat{\vy} \in \cB_{\cA} } \cL(\cF(\vx; \cM'), \widehat\vy)
\label{eq:loss}
\end{equation}

For clarification, we compare our framework with some recent works on billion-scale webly-supervised learning in Table~\ref{tab:difference}.
OmniSource is capable of dealing with web data from multiple sources. 
It is designed to help a specific task, treats webly-supervision as co-training across multiple data sources instead of pre-training, thus is much more data-efficient.
It is also noteworthy that our framework is orthogonal to webly-supervised pre-training~\cite{ghadiyaram2019large}.

\begin{table}
	\tiny
	\begin{center}
		\captionof{table}{\textbf{Difference to previous works.} The notions follow Sec.~\ref{subsec:formulation}: $ \cU $ is the unlabeled web data, $ \cD_\cT $ is the target dataset. $ \vert \cU \vert $, $ \vert \cD_\cA\vert $ denotes the scale of web data and filtered auxiliary dataset}
		\label{tab:difference}
	\begin{tabular}{l|l|l|l}
		&  Webly-supervised pretrain~\cite{mahajan2018exploring,ghadiyaram2019large} &  Web-scale semi-supervised~\cite{yalniz2019billion}  & OmniSource (Ours) \\
		\hline
		\multirow{5}{*}{\rotatebox[origin=c]{90}{Procedure}} & 1. Train a model $ \cM $ on $ \cU $. & 1. Train a model $ \cM $ on $ \cD_\cT $.                 &  1. Train one (or more) model $ \cM $ on $ \cD_\cT $. \\
		& 2. Fine-tune $ \cM $ on $ \cD_\cT $. & 2. Run $ \cM $ on $ \cU $ to pseudo-labeled $ \widehat{\cD} $. & 2. Run $ \cM $ on $\bigcup_i \cU_i $ to pseudo-labeled $ \bigcup_i{\widehat{\cD}_i} $. \\
		&                                      & 3. Train a student model $ \cM' $ on $ \widehat{\cD} $.   & (Samples under certain threshold are dropped.) \\
		&                                      & 4. Fine-tune $ \cM' $ on $ \cD_\cT $.                 & 3. Apply transforms $ \cT_i: \widehat{\cD}_i \rightarrow \cD_{\cA,i} $. \\
		& & & 4. Train model $ \cM' $ (or $ \cM $) on \bm{$ \cD_\cT \cup \cD_\cA $}. \\
		\hline
		\multirow{1}{*}{\rotatebox[origin=c]{90}{$ \vert \cU \vert $}} & 3.5B images \textbf{or} 65M videos & 1B images \textbf{or} 65M videos & $ \vert \cU \vert $: 13M images \textbf{and} 1.4M videos (\textbf{0.4\%}$\sim$\textbf{2\%}) \\ & & & $ \vert \cD_\cA \vert $: 3.5M images \textbf{and} 0.8M videos (\textbf{0.1\%}$\sim$\textbf{1\%})\\ \hline
	\end{tabular}
	\end{center}
	
\end{table}

\subsection{Task-specific data collection}
\label{subsec:data_collection}

We use class names as keywords for data \textbf{crawling}, with no extra query expansion. For tag-based system like Instagram, we use automatic permutation and stemming\footnote{For example, ``beekeeping'' can be transformed to ``beekeep'', and ``keeping bee''.} to generate tags.
We crawl web data from various sources, including search engine, social media and video sharing platform.
Because Google restricts the number of results for each query, we conduct multiple queries, each of which is restricted by a specific period of time.
Comparing with previous works ~\cite{mahajan2018exploring,ghadiyaram2019large} which rely on large-scale web data with hashtags, our task-specific collection uses keywords highly correlated with labels, making the supervision stronger.
Moreover, it reduces the required amount of web data by 2 orders of magnitude (\eg from 65M to 0.5M videos on Instagram).

After data collection, we first remove invalid or corrupted data. Since web data may contain samples very similar to validation data, data \textbf{de-duplication} is essential for a fair comparison. 
We perform content-based data de-duplication based on feature similarity. 
First, we extract frame-level features using an ImageNet-pretrained ResNet50.
Then, we calculate the cosine similarity of features between the web data and target dataset and perform pairwise comparison after whitening. 
The average similarity among different crops of the same frame is used as the threshold.
Similarity above it indicates suspicious duplicates.
For \texttt{Kinetics-400}, we filter out 4,000 web images (out of 3.5M, 0.1\%) and 400 web videos (out of 0.5M, 0.1\%). 
We manually inspect a subset of them and find that less than 10\% are real duplicates.

\subsection{Teacher filtering}
\label{subsec:teacher_filter}
Data crawled from the web are inevitably noisy.
Directly using collected web data for joint training leads to a significant performance drop (over 3\%).
To prevent irrelevant data from polluting the training set, we first train a teacher network $ \cM $ on the target dataset and discard those web data with low confidence scores.
For web images, we observe performance deterioration when deflating 3D teachers to 2D and therefore only use 2D teachers.
For web videos, we find both applicable and 3D teachers outperform 2D counterparts consistently.

\begin{figure}
	\begin{center}
		\includegraphics[width=.8\linewidth]{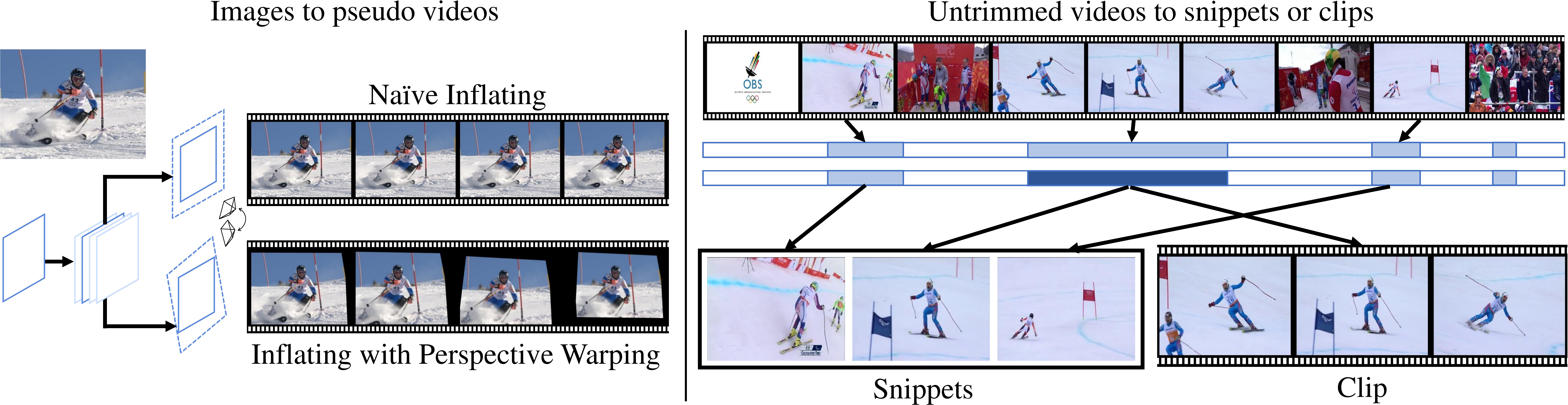}
	\end{center}
	\captionof{figure}{\textbf{Transformations.} Left: Inflating images to clips, by replicating or inflating with perspective warping; Right: Extracting segments or clips from untrimmed videos, guided by confidence scores}
	\label{fig:transform}
\end{figure}

\subsection{Transforming to the target domain}
\label{subsec:transform}
\noindent\textbf{Web Images.}
To prepare web images for video recognition training, we devise several ways to transform images into pseudo videos.
The first na\"ive way is to replicate the image $n$ times to form an $n$-frame clip. However, such clips may not be optimal since there is a visible gap between static clips and natural videos which visually change over time.
Therefore, we propose to generate video clips from static images by viewing them with a moving camera.
Given an image $ I $, under the standard perspective projection model~\cite{forsyth2002computer}, an image with another perspective $ \widetilde{I} $ can be generated by a homographic transform $ \cH $ which is induced by a homographic matrix $ \mH \in \Rbb^{3\times3} $, \ie, $ \widetilde{I} = \cH(I) = \cF(I; \mH) $.
To generate a clip $ J = \{ J_1, \cdots, J_N \}$ from $ I $, starting from $J_1 = I$, we have
\begin{equation}
	J_{i} = \cH_i(J_{i-1}) = (\cH_i \circ \cH_{i-1} \circ \cdots \circ \cH_{1})(I)
	\label{eq:warp}
\end{equation}
Each matrix $ \mH_i $ is randomly sampled from a multivariate Gaussian distribution $ \cN(\mu, \Sigma) $, while the parameters $ \mu $ and  $ \Sigma $ are estimated using maximum likelihood estimation on the original video source. 
Once we get pseudo videos, we can leverage web images for joint training with trimmed video datasets.

\noindent\textbf{Untrimmed Videos.}
Untrimmed videos form an important part of web data.
To exploit web untrimmed videos for video recognition, we adopt different transformations respectively for 2D and 3D architectures.

For 2D TSN, \emph{snippets} sparsely sampled from the entire video are used as input. 
We first extract frames from the entire video at a low frame rate (1 FPS).
A 2D teacher is used to get the confidence score of each frame, which also divides frames into positive ones and negative ones.
In practice, we find that only using positive frames to construct snippets is a sub-optimal choice.
Instead, combining negative frames and positive frames can form harder examples, results in better recognition performance.
In our experiments, we use 1 positive frame and 2 negative frames to construct a 3-snippet input.

For 3D ConvNets, video \emph{clips} (densely sampled continuous frames) are used as input. 
We first cut untrimmed videos into 10-second clips, then use a 3D teacher to obtain confidence scores.
Only positive clips are used for joint training.

\begin{figure}
	\begin{center}
		\includegraphics[width=.8\linewidth]{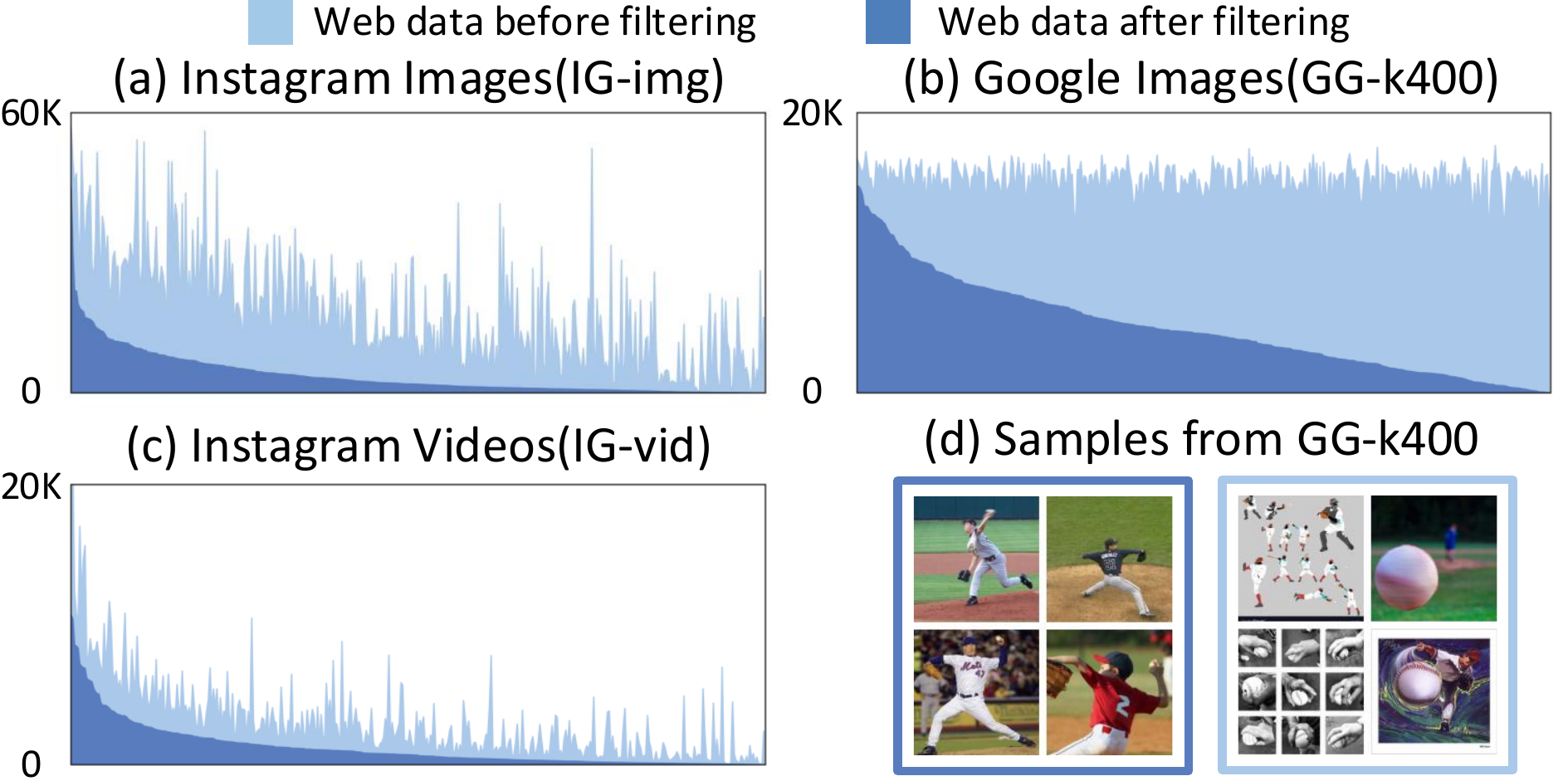}
	\end{center}
	\captionof{figure}{\textbf{Web Data Distribution.} The inter-class distribution of three web datasets is visualized in (a,b,c), both before and after filtering. (d) gives out samples of filtered out images (cyan) and remained images (blue) for \texttt{GG-K400}. Teacher filtering successfully filters out lots of negative examples while making inter-class distribution more uneven}
	\label{fig:web_dist}
\end{figure}

\subsection{Joint training}
\label{subsec:joint_train}
Once web data are filtered and transformed into the same format of that in the target dataset $ \cD_{\cT} $, we construct an \emph{auxiliary} dataset $ \cD_{\cA} $.
A network can then be trained with both $ \cD_\cT $ and $ \cD_\cA $ using sum of cross-entropy loss in Eq.~\ref{eq:loss}.
As shown in Fig.~\ref{fig:web_dist}, web data across classes are extremely unbalanced, especially after teacher filtering.
Also there exists potential domain gap between $ \cD_\cT $ and $ \cD_\cA $. 
To mitigate these issues, we enumerate several good practices as follows.

\noindent\textbf{Balance between target and auxiliary mini-batches.}
Since the auxiliary dataset may be much larger than the target dataset and the domain gap may occur, the data ratio between target and auxiliary mini-batches is crucial for the final performance.
Empirically, $\vert \cB_{\cT} \vert : \vert \cB_{\cA} \vert = 2:1 \sim 1:1 $ works reasonably well.

\noindent\textbf{Resampling strategy.}
Web data are extremely unbalanced, especially after teacher filtering (see Fig~\ref{fig:web_dist}).
To alleviate this, we explore several sampling policies:
(1) sampling from a clipped distribution: classes whose samples exceeds threshold $ N_c $ are clipped;
(2) sampling from distribution modified by a power law: the probability of choosing class with $ N $ samples is proportional to $ N ^ p $($ p\in (0, 1)$).
We find that (2) parameterized by $p = 0.2$ is generally a better practice.

\noindent\textbf{Cross-dataset \emph{mixup}.}
Mixup~\cite{zhang2017mixup} is a widely used strategy in image recognition.
It uses convex combinations of pairs of examples and their labels for training, thus improving the generalization of deep neural networks.
We find that technique also works for video recognition.
When training teacher networks on $ \cD_\cT $ only, we use the linear combination of two clip-label pairs as training data, termed as \emph{intra-dataset} mixup.
When both target and auxiliary datasets are used, the two pairs are samples randomly chosen from both datasets, termed as \emph{cross-dataset} mixup. 
Mixup works fairly well when networks are trained from scratch.
For fine-tuning, the performance gain is less noticeable.

\section{Datasets}
In this section, we introduce the datasets on which experiments will be conducted.
Then we go through different sources from which web data are collected.

\subsection{Target datasets}
\noindent\textbf{Kinetics-400}
The Kinetics dataset~\cite{carreira2017quo} is one of the largest video datasets.
We use the version released in 2017 which contains 400 classes and each category has more than 400 videos.
In total, it has around 240K, 19K, and 38K videos for training, validation and testing subset respectively.
In each video, a 10-second clip is annotated and assigned a label.
These 10-second clips constitute the data source for the default supervised learning setting, which we refer to \textbf{\texttt{K400-tr}}.
The rest part of training videos is used to mimic untrimmed videos sourced from the Internet which we refer to \textbf{\texttt{K400-untr}}.

\noindent\textbf{Youtube-car}
Youtube-car~\cite{zhu2018fine} is a fine-grained video dataset with 10K training and 5K testing videos of 196 types of cars. 
The videos are untrimmed, last several minutes.
Following ~\cite{zhu2018fine}, the frames are extracted from videos at 4 FPS.

\noindent\textbf{UCF101}
UCF101~\cite{soomro2012ucf101} is a small scale video recognition dataset, which has 101 classes and each class has around 100 videos. 
We use the official split-1 in our experiments, which has about 10K and 3.6K videos for training and testing.

\subsection{Web sources}
\label{sec:web_data}
We collect web images and videos from various sources including search engines, social medias and video sharing platforms.

\noindent\textbf{GoogleImage}
GoogleImage is a search engine based web data source for Kinetics-400, Youtube-car and UCF101.
We query each class name in the target dataset on Google to get related web images.
We crawl 6M, 70K, 200K URLs for Kinetics-400, Youtube-car and UCF101 respectively.
After data cleaning and teacher filtering, about 2M, 50K, 100K images are used for training on these three datasets.
We denote the three datasets as \textbf{\texttt{GG-k400}}, \textbf{\texttt{GG-car}}, and \textbf{\texttt{GG-UCF}} respectively.

\noindent\textbf{Instagram}
Instagram is a social media based web data source for Kinetics-400.
It consists of InstagramImage and InstagramVideo.
We generate several tags for each class in Kinetics-400, resulting in 1,479 tags and 8.7M URLs.
After removing corrupted data and teacher filtering, about 1.5M images and 500K videos are used for joint training, denoted as \textbf{\texttt{IG-img}} and \textbf{\texttt{IG-vid}}.
As shown in Fig~\ref{fig:web_dist}, \texttt{IG-img} is significantly unbalanced after teacher filtering.
Therefore, in the coming experiments, \texttt{IG-img} is used in combination with \texttt{GG-k400}.

\noindent\textbf{YoutubeVideo}
YoutubeVideo is a video sharing platform based web data source for Youtube-car.
We crawl 28K videos from youtube by querying class names. After de-duplicating (remove videos in the original Youtube-car dataset) and teacher filtering, 17K videos remain, which we denote as \textbf{\texttt{YT-car-17k}}.

\section{Experiments}
\subsection{Video architectures}
We mainly study two families of video classification architectures, namely Temporal Segment Networks~\cite{wang2018temporal} and 3D ConvNets~\cite{carreira2017quo}, to verify the effectiveness of our design. Unless specified, we use ImageNet-pretrained models for initialization. We conduct all experiments using MMAction~\cite{zhao2019mmaction}. 

\noindent\textbf{2D TSN}
Different from the original setting in~\cite{wang2018temporal}, we choose ResNet-50~\cite{he2016deep} to be the backbone, unless otherwise specified.
The number of segments is set to be 3 for Kinetics/UCF-101 and 4 for Youtube-car, respectively.

\noindent\textbf{3D ConvNets} For 3D ConvNet, we use the SlowOnly architecture proposed in~\cite{feichtenhofer2019slowfast} in most of our experiments. 
It takes 64 consecutive frames as a video clip and sparsely samples 4/8 frames to form the network input.
Different initialization strategies are explored, including training from scratch and fine-tuning from a pre-trained model.
Besides, more advanced architecture like Channel Separable Network~\cite{tran2019video} and more powerful pre-training (IG-65M~\cite{ghadiyaram2019large}) is also explored.

\subsection{Verifying the efficacy of OmniSource}

We verify our framework's efficacy by examining several questions.

\noindent\textbf{Why do we need teacher filtering and are search results good enough?}
Some may question the necessity of a teacher network for filtering
under the impression that a modern search engine might have internally utilized a visual recognition model, possibly trained on massively annotated data, to help generate the search results.
However, we argue that web data are inherently noisy and we observe nearly half of the returned results are irrelevant.
More quantitatively, 70\% - 80\% of the web data are rejected by the teacher.
On the other hand, we conduct an experiment without teacher filtering.
Directly using collected web data for joint training leads to a significant (over 3\%) performance drop on TSN.
This reveals that teacher filtering is necessary to help retain the useful information from the crawled web data while eliminating the useless.

\noindent\textbf{Does every data source contribute?}
We explore the contribution of different source types: images, trimmed videos and untrimmed videos.
For each data source, we construct auxiliary dataset and use it for joint training with \texttt{K400-tr}.
Results in Table~\ref{tab:omnisource} reveal that every source contributes to improving accuracy on the target task. 
When combined, the performance is further improved.

\begin{table}
	\scriptsize
	\begin{center}
		\captionof{table}{\textbf{Every source contributes.} We find that each source contributes to the target task. 
			With all sources combined(we intuitively set the ratio as: \texttt{K400-tr}  : \texttt{Web-img} : \texttt{IG-vid} : \texttt{K400-untr} = 2: 1: 1: 1), the improvement can be more considerable.
			The conclusion holds for both 2D TSN and 3D ConvNets (Format: Top-1 Acc/ Top-5 Acc)}
		\label{tab:omnisource}
		\begin{tabular}{c||c||c|c|c|c||c}
			\hline \hline
			
			Arch/Dataset  & \texttt{K400-tr}  & +\texttt{GG-k400} & +\texttt{GG\&IG-img} & +\texttt{IG-vid} & +\texttt{K400-untr} & + All \\
			\hline \hline
			\begin{tabular}[c]{@{}c@{}}TSN-3seg\\ R50 \end{tabular}  & 70.6/89.4 & 71.5/89.5 & 72.0/90.0 & 72.0/90.3 & 71.7/89.6  & 73.6/91.0 \\ \hline
			\begin{tabular}[c]{@{}c@{}}SlowOnly\\ 4x16,R50 \end{tabular} & 73.8/90.9 & 74.5/91.4 & 75.2/91.6 & 75.2/91.7 & 74.5/91.1  & 76.6/92.5 \\ 
			\hline \hline
		\end{tabular}
	\end{center}
\end{table}

For images, when the combination of \texttt{GG-k400} and \texttt{IG-img} is used, the Top-1 accuracy increases around 1.4\%.
For trimmed videos, we focus on \texttt{IG-vid}.
Although being extremely unbalanced, \texttt{IG-vid} still improves Top-1 accuracy by over 1.0\% in all settings.
For untrimmed videos, we use the untrimmed version of Kinetics-400 (\texttt{K400-untr}) as the video source and find it also works well.

\noindent\textbf{Do multiple sources outperform a single source? } 
Seeing that web data from multiple sources can jointly contribute to the target dataset, we wonder if multiple sources are still better than a single source with the same budget.
To verify this, we consider the case of training TSN on both \texttt{K400-tr} and $ \cD_\cA =\texttt{GG-k400}+\texttt{IG-img} $.
We fix the scale of auxilary dataset to be that of \texttt{GG-k400} and vary the ratio between \texttt{GG-k400} and \texttt{IG-img} by replacing images from \texttt{GG-k400} with those in \texttt{IG-img}.
From Fig.~\ref{fig:multisource}, we observe an improvement of 0.3\% without increasing $ \vert \cD_\cA \vert $, indicating that multiple sources provide complementary information by introducing diversity.

\noindent\textbf{Does OmniSource work with different architectures?} 
We further conduct experiments on a wide range of architectures and obtain the results in Table~\ref{tab:improvement}. 
For TSN, we use EfficientNet-B4~\cite{tan2019efficientnet} instead as the backbone, on which OmniSource improves Top-1 accuracy by 1.9\%.
For 3D-ConvNets, we conduct experiments on the SlowOnly-8x8-ResNet101 baseline, which takes longer input and has a larger backbone.
Our framework also works well in this case, improving the Top-1 accuracy from 76.3\% to \textbf{80.4\%} when training from scratch, from 76.8\% to \textbf{80.5\%} with ImageNet pretraining. 
The improvement on larger networks is higher, suggesting that deeper networks are more prone to suffering from the scarcity of video data and OmniSource can alleviate this.

\noindent
\begin{minipage}{\linewidth}
	\begin{minipage}{0.73\linewidth}
		\begin{center}
			\scriptsize
			\begin{tabular}{c|c|c|c|c|c}
				\hline \hline
				Arch          & Backbone     & Pretrain & w/o. Omni & w/. Omni & $ \Delta $ \\ \hline \hline
				TSN-3seg      & ResNet50     & ImageNet & 70.6 / 89.4  & 73.6 / 91.0  & +3.0 / +1.6  \\ \hline
				TSN-3seg      & ResNet50     & IG-1B    & 73.1 / 90.4  & 75.7 / 91.9  & +2.6 / +1.5  \\ \hline
				TSN-3seg      & Efficient-b4 & ImageNet & 73.3 / 91.0  & 75.2 / 92.0  & +1.9 / +1.0  \\ \hline
				SlowOnly-4x16 & ResNet50     &  -  & 72.9 / 90.9  & 76.8 / 92.5  & +3.9 / +1.6  \\ \hline
				SlowOnly-4x16 & ResNet50     & ImageNet & 73.8 / 90.9  & 76.6 / 92.5 & +2.8 / +1.6   \\ \hline
				SlowOnly-8x8  & ResNet101    &  -  &     76.3 / 92.6    &   80.4 / 94.4  &  +4.1 / +1.8  \\ \hline
				SlowOnly-8x8  & ResNet101    & ImageNet & 76.8 / 92.8  &  80.5 / 94.4 & +3.7 / +1.6 \\ \hline
				irCSN-32x2    & irCSN-152    & IG-65M   & 82.6 / 95.3  & 83.6 / 96.0 & +1.0 / +0.7 \\ \hline \hline
			\end{tabular}
		\end{center}
		\captionof{table}{\textbf{Improvement under various experiment configurations.}
			OmniSource is extensively tested on various architectures with various pretraining strategies. 
			The improvement is significant in \textbf{ALL} tested choices.
			Even for the SOTA setting, which uses 65M web videos for pretraining, OmniSource still improves the Top-1 accuracy by 1.0\% (Format: Top-1 / Top-5 Acc)}
		\label{tab:improvement}
	\end{minipage}
	\hfill
	\begin{minipage}{.24\linewidth}
		\begin{center}
			\includegraphics[width=.85\linewidth]{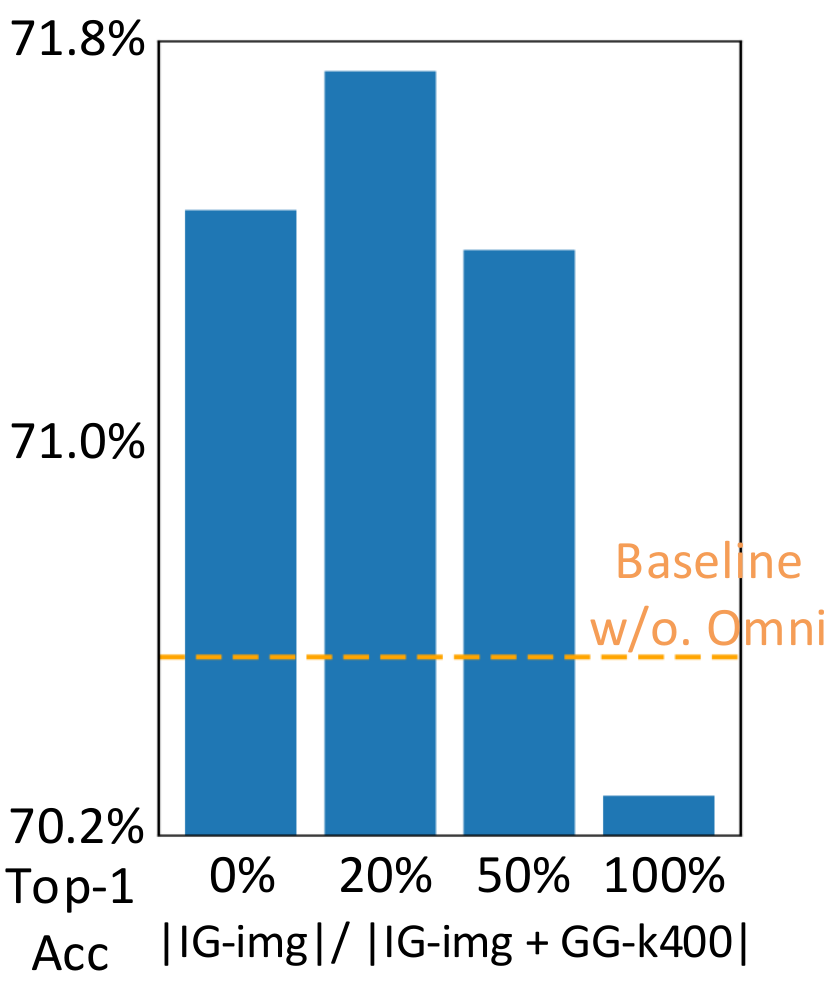}
		\end{center}
		\captionof{figure}{\textbf{Multi-source is better. }  Mixed sources lead to better performance with a constrained number of web images}
		\label{fig:multisource}
	\end{minipage}
	
\end{minipage}

\noindent\textbf{Is OmniSource compatible with different pre-training strategies?}
As discussed, OmniSource alleviates the data-hungry issue by utilizing auxiliary data.
One natural question is: how does it perform when training 3D networks from scratch?
Can we simply drop ImageNet pretraining in pursuit of a more straightforward training policy?
Indeed, we find that OmniSource works fairly well under this setting and interestingly the performance gain is more significant than fine-tuning.
For example, SlowOnly-(4x16, R50) increases the Top-1 accuracy by 3.9\% when training from scratch while fine-tuning only increases by 2.8\%.
The model trained from scratch beats the fine-tuned counterpart by 0.2\% with OmniSource though being 0.9\% lower with only \texttt{K400-tr}.
Similar results can be observed for SlowOnly-(8x8, R101). 
With large-scale webly supervised pretraining, OmniSource still leads to significant performance improvement (+2.6\% Top-1 for TSN-3seg-R50, +1.0\% Top-1 for irCSN-32x2).

\noindent
\begin{minipage}{\linewidth}
	\begin{minipage}[b]{.6\linewidth}
		\scriptsize
		\begin{center}
			\begin{tabular}{c||c|c||c|c}
				\hline \hline
				\multirow{2}{*}{Arch}                                                & \multicolumn{2}{c||}{UCF101-split1} & \multicolumn{2}{c}{HMDB51-split1} \\ \cline{2-5} 
				& w.o. Omni         & w/. Omni        & w/o. Omni         & w. Omni        \\ \hline \hline
				\begin{tabular}[c]{@{}c@{}}TSN-3seg\\ R50{[}FT{]}\end{tabular}       & 91.5              & 93.3           & 63.5              & 65.9           \\ \hline
				\begin{tabular}[c]{@{}c@{}}SlowOnly\\ 4x16, R50{[}FT{]}\end{tabular} & 94.7              & 96.0           & 69.4              & 70.7           \\ \hline
				\begin{tabular}[c]{@{}c@{}}SlowOnly\\ 4x16, R50{[}SC{]}\end{tabular} & 94.1              & 96.0           & 65.8              & 71.0           \\ \hline \hline
			\end{tabular}
		\end{center}
		\captionof{table}{\textbf{OmniSource features transfer well.} We finetune on UCF101 and HMDB51 with \texttt{K400-tr} pretrained weight. Pretraining with OmniSource improves the performance significantly}
		\label{tab:transfer}
	\end{minipage}
	\hfill
	\begin{minipage}[b]{.38\linewidth}
		\begin{center}
			\includegraphics[width=.9\linewidth]{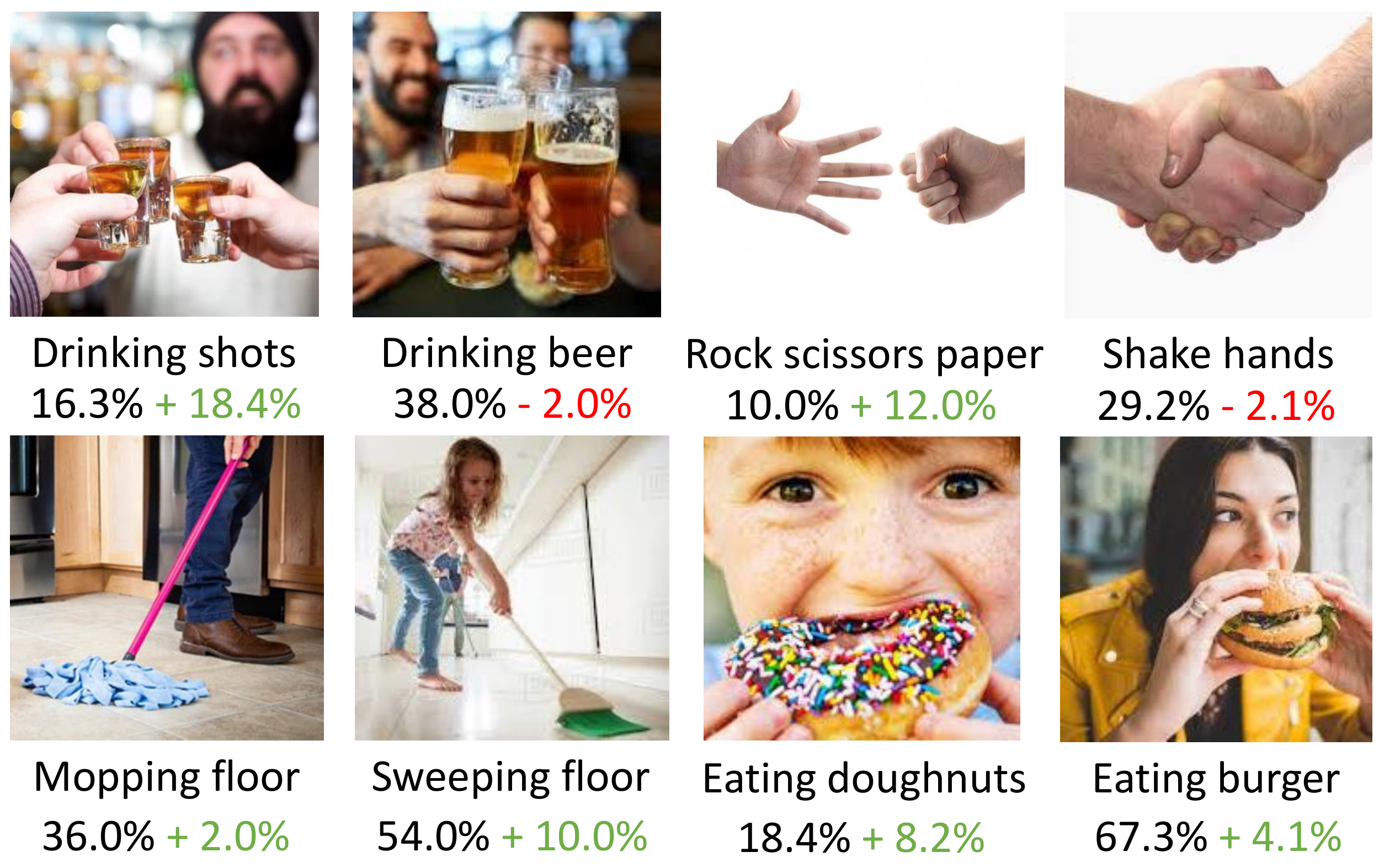}
		\end{center}
		\captionof{figure}{\textbf{Confusing pairs improved by OmniSource.}
		The original accuracy and change are denoted in black and in color}
		\label{fig:improvement}
	\end{minipage}
\end{minipage}

\noindent\textbf{Do features learned by OmniSource transfer to other tasks?}
Although OmniSource is designed for a target video recognition task, the learned features also transfer well to other video recognition tasks. 
To evaluate the transfer capability, we finetune the learned model on two relatively smaller datasets:
UCF101~\cite{soomro2012ucf101} and HMDB51~\cite{kuehne2011hmdb}. 
Table~\ref{tab:transfer} indicates that on both benchmarks, pretraining with OmniSource leads to significant performance improvements.
Following standard evaluation protocol, SlowOnly-8x8-R101 achieves 97.3\% Top-1 accuracy on UCF101, 79.0\% Top-1 accuracy on HMDB51 with RGB input.
When combined with optical flow, it achieves \textbf{98.6\%} and \textbf{83.8\%} Top-1 accuracy on UCF101 and HMDB51, which is the new state-of-the-art. 
More results on transfer learning are provided in the supplementary material.

\noindent\textbf{Does OmniSource work in different target domains?}
Our framework is also effective and efficient in various domains.
For a fine-grained recognition benchmark called \textbf{Youtube-car}, we collect 50K web images (\texttt{GG-car}) and 17K web videos (\texttt{YT-car-17k}) for training. 
Table~\ref{tab:youtube-car} shows that the performance gain is significant: 5\% in both Top-1 accuracy and mAP. 
On \textbf{UCF-101}, we train a two-stream TSN network with BNInception as the backbone. 
The RGB stream is trained either with or without \texttt{GG-UCF}. 
The results are listed in Table~\ref{tab:ucf-101}. 
The Top-1 accuracy of the RGB stream improves by 2.7\%.
When fused with the flow stream, there is still an improvement of 1.1\%.

\noindent\textbf{Where does the performance gain come from?}
To find out why web data help, we delve deeper into the collected web dataset and analyze the improvement on individual classes.
We choose TSN-3seg-R50 trained either with or without \texttt{GG-k400}, where the improvement is 0.9\% on average.
We mainly focus on the confusion pairs that web images can improve.
We define the \emph{confusion score} of a class pair as
$ s_{ij} = (n_{ij} + n_{ji}) / (n_{ij} + n_{ji} + n_{ii} + n_{jj}) $, 
where $n_{ij}$ denotes the number of images whose ground-truth are class $i$ while being recognized as class $j$.
Lower confusion score denotes better discriminating power between the two classes.
We visualize some confusing pairs in Fig~\ref{fig:improvement}.
We find the improvement can be mainly attributed to two reasons:
(1) Web data usually focus on key objects of action.
For example, we find that in those pairs with the largest confusion score reduction, there exist pairs like ``drinking beer'' \vs ``drinking shots'', and ``eating hotdog'' \vs ``eating chips''. Training with web data leads to better object recognition ability in some confusing cases.
(2) Web data usually include discriminative poses, especially for those actions which last for a short time.
For example, ``rock scissors paper'' \vs ``shaking hands'' has the second-largest confusion score reduction.
Other examples including ``sniffing''-``headbutting'', ``break dancing''-``robot dancing'', etc.

\noindent
\begin{minipage}{\linewidth}
	\begin{minipage}{.3\linewidth}
		\centering
		\captionof{table}{Youtube-car}
		\label{tab:youtube-car}
		\scriptsize
		\begin{tabular}{ccc}
			\hline
			Setting    & Top-1 &  mAP\\
			\hline
			Baseline   & 77.05 &  71.95 \\
			\hline
			+\texttt{GG-car}  &  80.96 &  77.05 \\
			\hline
			+\texttt{YT-car-17k} &  81.68 & 78.61 \\
			\hline
			+\texttt{[GG-]}+\texttt{[YT-]}  & 81.95  & 78.67 \\
			\hline
		\end{tabular}
		\centering
		\captionof{table}{UCF-101}
		\label{tab:ucf-101}
		\begin{tabular}{ccc}
			\hline
			Setting  & + Flow  & Top-1 \\
			\hline
			Baseline   &  &  86.04 \\
			\hline
			+ \texttt{GG-UCF} &  & 88.74  \\
			\hline
			Baseline & \cmark & 93.47 \\
			\hline
			+ \texttt{GG-UCF} & \cmark  & 94.58\\
			\hline
		\end{tabular}
	\end{minipage} 
	\hfill
	\begin{minipage}{.68\linewidth}
		\scriptsize
		\begin{center}
			\captionof{table}{Comparisons with Kinetics-400 state-of-the-art }
			\label{tab:sota}
			\begin{tabular}{ccccc}
				\hline
				Method & backbone & pretrain &  Top-1  & Top-5 \\ \hline \hline
				TSN-7seg~\cite{wang2018temporal} & Inception-v3 & ImageNet & 73.9 & 91.1 \\ \hline
				TSM-8seg~\cite{lin2019tsm} & ResNet50 & ImageNet & 72.8 & N/A \\ \hline 
				TSN-3seg (\textbf{Ours}) & ResNet50 & ImageNet & 73.6 & 91.0 \\  \hline
				TSN-3seg (\textbf{Ours}) & Efficient-b4 & ImageNet & \textbf{75.2} & \textbf{92.0} \\  \hline \hline
				SlowOnly-8x8~\cite{feichtenhofer2019slowfast} & ResNet101 & - & 75.9 & N/A \\ \hline
				SlowFast-8x8~\cite{feichtenhofer2019slowfast} & ResNet101 & - & 77.9 & 93.2 \\ \hline
				SlowOnly-8x8 (\textbf{Ours}) & ResNet101 & - & \textbf{80.4} & \textbf{94.4} \\ \hline
				I3D-64x1~\cite{carreira2017quo} & Inception-V1 & ImageNet & 72.1 & 90.3 \\ \hline
				NL-128x1~\cite{wang2018non} & ResNet101 & ImageNet & 77.7 & 93.3 \\ \hline
				SlowFast-8x8~\cite{feichtenhofer2019slowfast} & ResNet101 & ImageNet &  77.9 & 93.2 \\ \hline
				LGD-3D (RGB)~\cite{qiu2019learning} & ResNet101 & ImageNet & 79.4 & 94.4 \\ \hline
				STDFB~\cite{martinez2019action} & ResNet152 & ImageNet & 78.8 & 93.6 \\ \hline
				SlowOnly-8x8 (\textbf{Ours}) & ResNet101 & ImageNet &  \textbf{80.5} & \textbf{94.4} \\ \hline
				irCSN-32x2~\cite{ghadiyaram2019large} & irCSN-152 & IG-65M & 82.6 & 95.3 \\ \hline
				irCSN-32x2 (\textbf{Ours}) & irCSN-152 & IG-65M & \textbf{83.6} & \textbf{96.0} \\ \hline \hline
			\end{tabular}
		\end{center}
	\end{minipage}
\end{minipage}

\subsection{Comparisons with state-of-the-art}
In Table~\ref{tab:sota}, we compare OmniSource with current state-of-the-art on Kinetics-400.
For 2D ConvNets, we obtain competitive performance with fewer segments and lighter backbones.
For 3D ConvNets, considerable improvement is achieved for all pre-training settings with OmniSource applied.
With IG-65M pre-trained irCSN-152,
OmniSource achieves \textbf{83.6\%} Top-1 accuracy, an absolute improvement of 1.0\% with only 1.2\% relatively more data, establishing a new record.

\noindent
\begin{minipage}{\linewidth}
	\scriptsize
	\begin{minipage}{.44\linewidth}
		\setlength{\tabcolsep}{2pt}
		\centering
		\captionof{table}{Different ways to transform images into video clips. Still inflation is a strong baseline, while agnostic perspective warping performs best}
		\label{tab:inflation}
		\begin{tabular}{ccc}
			\hline
			Inflation & Top-1 & Top-5 \\ \hline
			N/A &  73.8 & 90.9 \\ \hline
			replication (still) &  74.1 & 91.2 \\ \hline
			translation (random) &  73.7 & 90.9 \\ \hline
			translation (constant) &  73.8 & 90.8 \\ \hline
			perspective warp {[}spec{]} &  74.4 & 91.3 \\ \hline
			perspective warp {[}agno{]} &  74.5 & 91.4 \\ \hline
		\end{tabular}
	\end{minipage}%
	\hfill
	\begin{minipage}{.52\linewidth}
		\setlength{\tabcolsep}{2pt}
		\centering
		\captionof{table}{Mixup technique can be beneficial to the model performance, both for intra- and cross-dataset cases. However, it works only when the model is trained from scratch }
		\label{tab:mixup}
		\begin{tabular}{ccccc}
			\hline
			Pretraining & w. mixup & w.GG-img  & Top-1 & Top-5  \\ \hline
			ImageNet &       &   & 73.8 & 90.9 \\ \hline
			ImageNet & \cmark&   & 73.6 & 91.1 \\ \hline
			None     & 	     &   & 72.9 & 90.9 \\ \hline
			None     & \cmark&   & 73.3 & 90.9 \\ \hline
			None     &       &  \cmark & 74.1 & 91.0 \\ \hline
			None     & \cmark&  \cmark & 74.4 & 91.4 \\ \hline
		\end{tabular}
	\end{minipage} 
\end{minipage}

\subsection{Validating the good practices in OmniSource}

We conduct several ablation experiments on techniques we introduced.
The target dataset is \texttt{K400-tr} and the auxiliary dataset is \texttt{GG-k400} unless specified.

\noindent\textbf{Transforming images to video clips.}
We compare different ways to transform web images into clips in Table~\ref{tab:inflation}.
Na\"ively replicating still image brings limited improvement (0.3\%).
We then apply translation with randomized or constant speed to form pseudo clips.
However, the performance deteriorates slightly, suggesting that translation cannot mimic the camera motion well.
Finally, we resort to perspective warping to hallucinate camera motion.
Estimating class-agnostic distribution parameters is slightly better,
suggesting that all videos might share similar camera motion statistics.

\noindent\textbf{Cross-Dataset mixup}
In Table~\ref{tab:mixup}, we find that mixup is effective for video recognition in both intra- and cross-dataset cases when the model is trained from scratch. 
The effect is unclear for fine-tuning.
In particular, mixup can lead to 0.4\% and 0.3\% Top-1 accuracy improvement for intra- and inter-dataset cases.

\noindent\textbf{Impact of teacher choice.}
Since both teacher and student networks can be 2D or 3D ConvNets, there are 4 possible combinations for teacher network choosing.
For images, deflating 3D ConvNets to 2D yields a dramatic performance drop.
Therefore, we do not use 3D ConvNet teachers for web images.
For videos, however, 3D ConvNets lead to better filtering results comparing to its 2D counterpart.  
To examine the effect of different teachers, we fix the student model to be a ResNet-50 and vary the choices of teacher models (ResNet-50, EfficientNet-b4, and the ensemble of ResNet-152 and EfficientNet-b4).
Consistent improvement is observed against the baseline (70.6\%).
The student accuracy increases when a better teacher network is used.
It also holds for 3D ConvNets on web videos.

\noindent\textbf{Effectiveness when labels are limited.}
To validate the effectiveness with limited labeled data, we construct 3 subsets of \texttt{K400-tr} with a proportion of 3\%, 10\%, and 30\% respectively.
We rerun the entire framework including data filtering with a weaker teacher.
The final results on the validation set of \texttt{K400-tr} is shown in Fig~\ref{fig:partial_kinetics}.
Our framework consistently improves the performance as the percentage of labeled videos varies.
Particularly, the gain is more significant when data are scarce, \eg a relative increase of over 30\% with 3\% labeled data.

\noindent
\begin{minipage}{\linewidth}
	\begin{minipage}{0.31\linewidth}
		\begin{center}
			\includegraphics[width=1\linewidth]{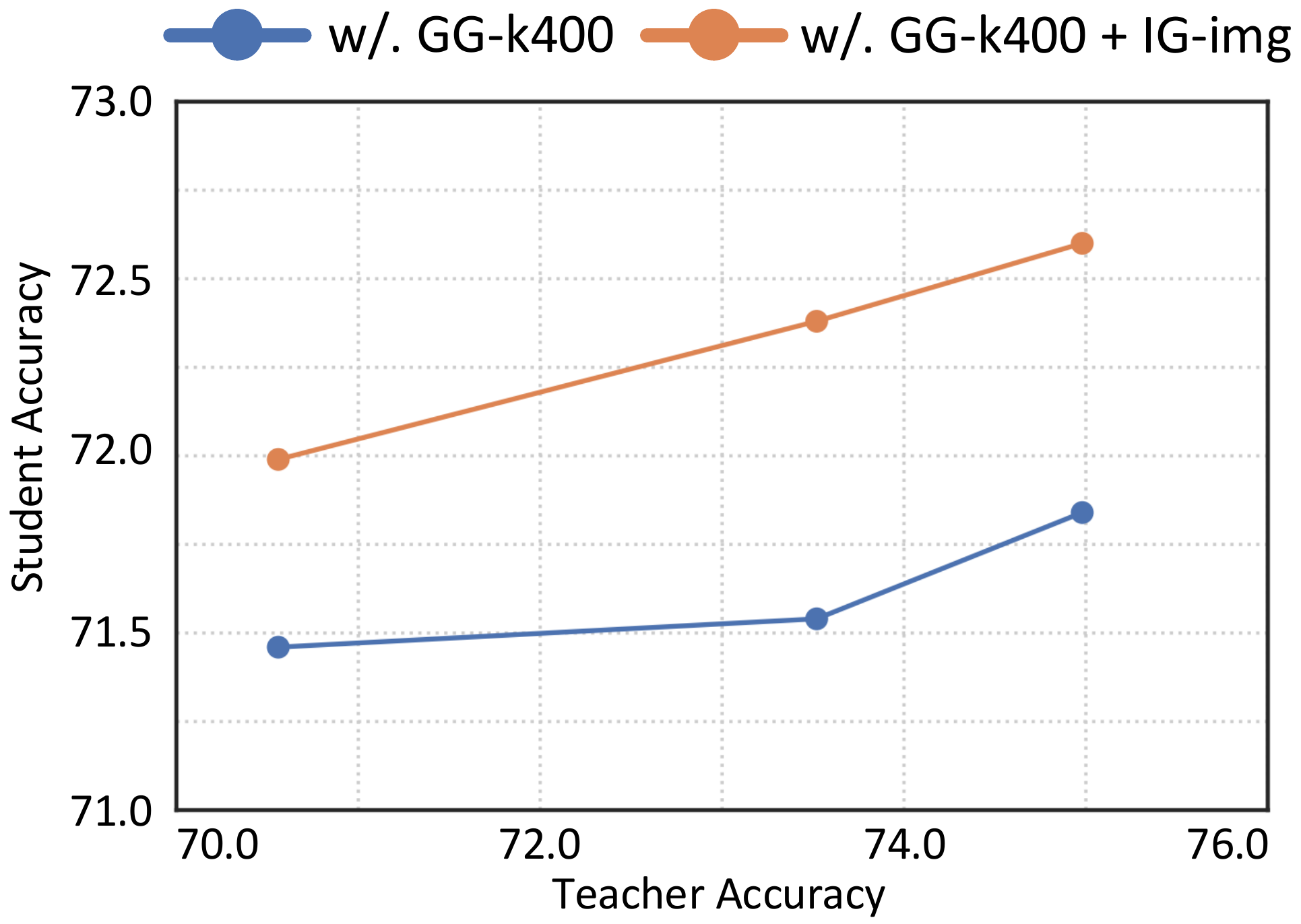}
		\end{center}
		\captionof{figure}{Better teachers leads to better students}
		\label{fig:teacher_quality}
	\end{minipage}
	\hfill
	\begin{minipage}{0.31\linewidth}
		\begin{center}
			\includegraphics[width=1\linewidth]{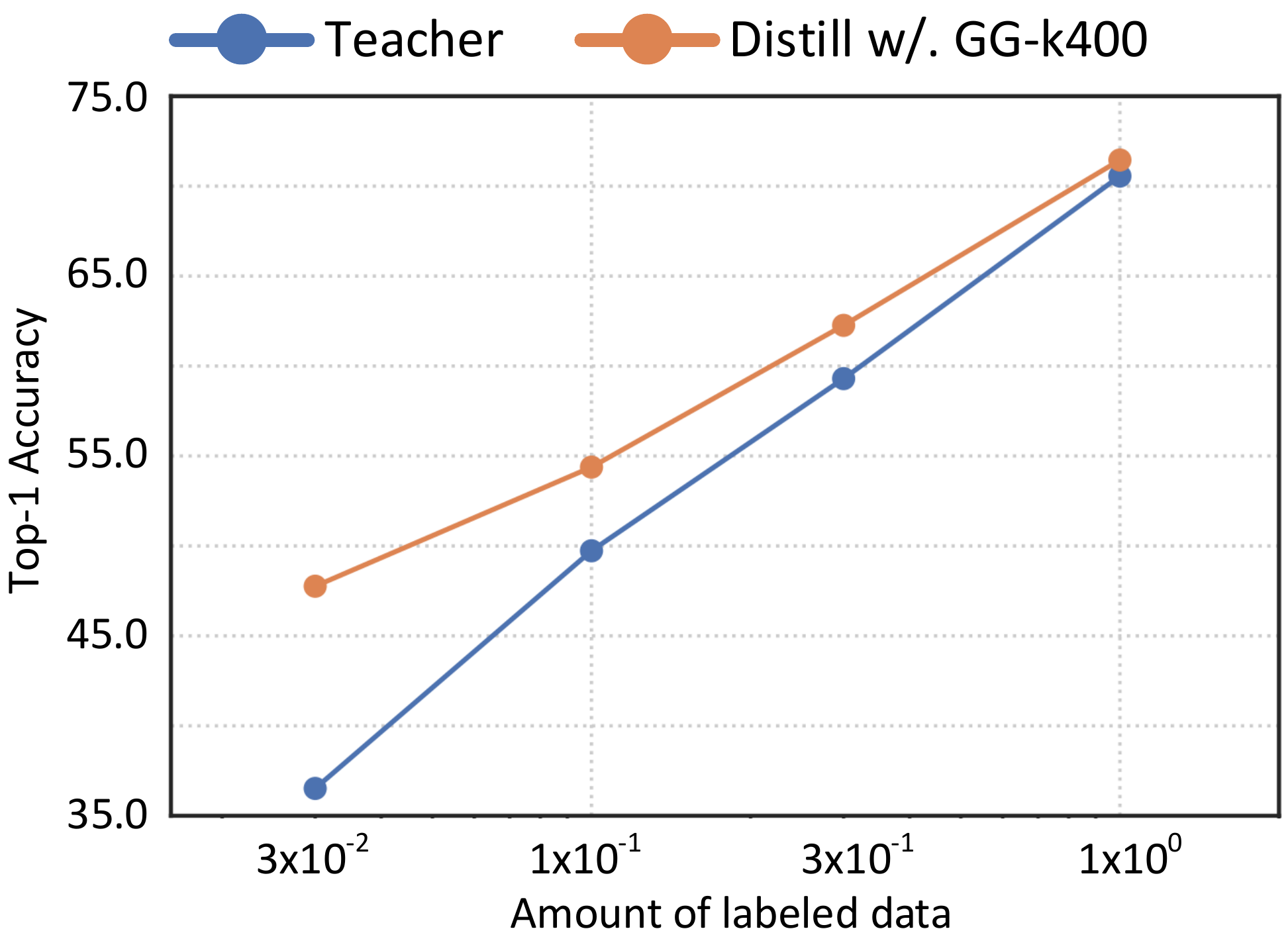}
		\end{center}
		\captionof{figure}{OmniSource on subsets of Kinetics}
		\label{fig:partial_kinetics}
	\end{minipage}
    \hfill
	\begin{minipage}{0.31\linewidth}
		\begin{center}
			\includegraphics[width=1\linewidth]{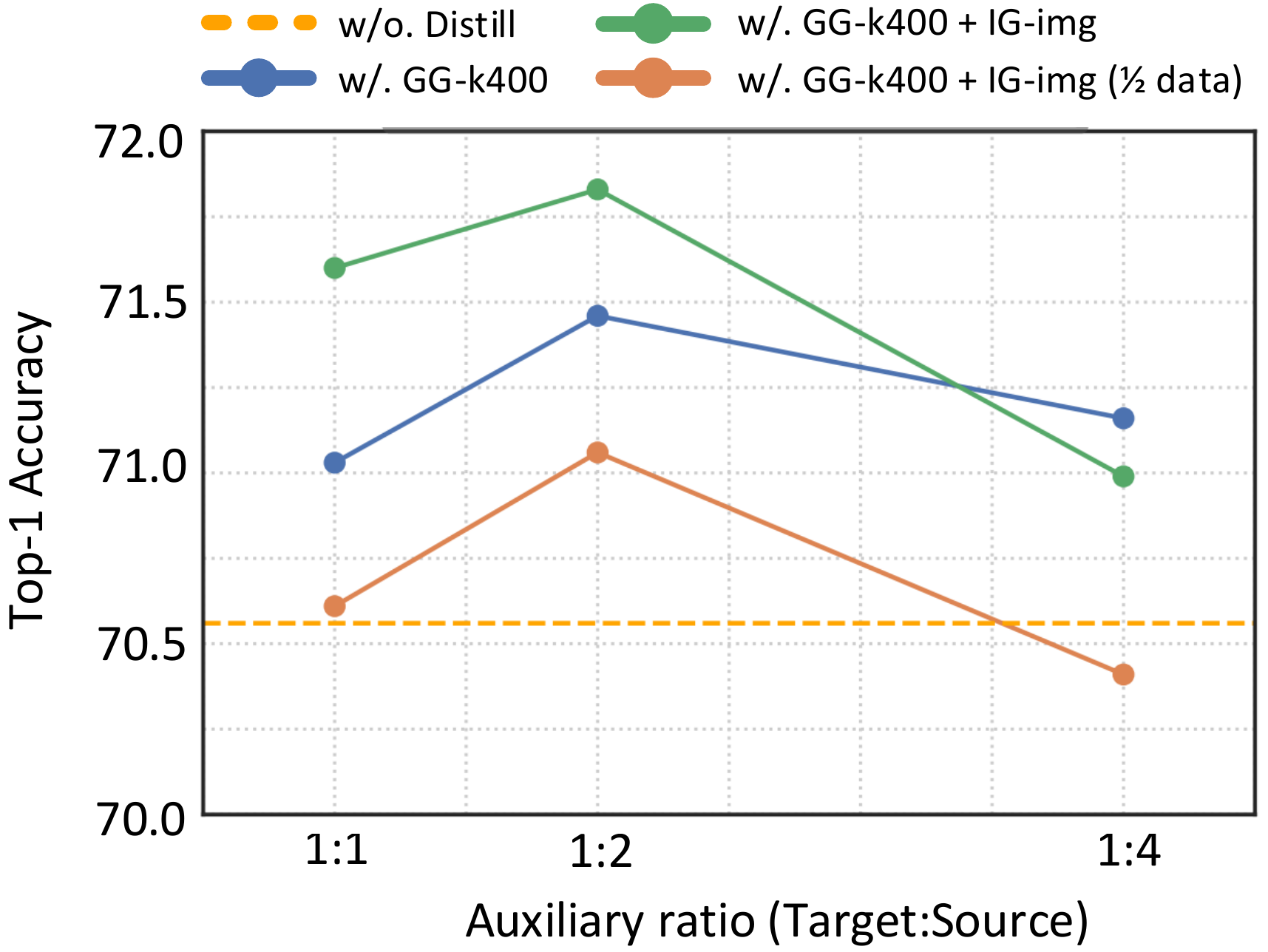}
		\end{center}
		\captionof{figure}{Accuracy with different ratios $ \vert \cB_\cT \vert : \vert \cB_\cA \vert $ }
		\label{fig:auxiliary_ratio}
	\end{minipage}
\end{minipage}

\begin{wraptable}{r}{.32\linewidth}	
	\begin{center}
		\scriptsize
		\caption{\textbf{Resampling strategies.} Simple resampling strategies lead to nontrivial improvement}
		\label{tab:resample}
		\setlength{\tabcolsep}{0.2pt}
		\begin{tabular}{cc}
			\hline
			Stategy              &  Top-1/5\\
			\hline
			None (original)      & 71.5/89.5 \\
			\hline
			Clipped ($ N_c = 5000 $)  &  71.9/90.0 \\
			\hline
			Power ($ \sim N^p, p=0.5 $) &  71.8/89.7 \\
			\hline
			Power ($ \sim N^p, p=0.2 $) &  72.0/90.0 \\
			\hline
		\end{tabular}
	\end{center}
	
\end{wraptable}

\noindent\textbf{Balancing between the target and auxiliary dataset.}
We tune the ratio between the batch size of the target dataset $ \vert \cB_{\cT} \vert $ and the auxiliary dataset $ \vert \cB_{\cA} \vert $ and obtain the accuracy on Fig~\ref{fig:auxiliary_ratio}.
We test 3 scenarios: (1) the original \texttt{GG-k400}, clarified in Sec~\ref{sec:web_data}; (2) \texttt{[GG+IG]-k400}, the union of \texttt{GG-k400} and \texttt{IG-img}; (3) \texttt{[GG+IG]-k400-half} which is the half of (2).
We observe that the performance gain is robust to the choice of $\vert \cB_\cT \vert / \vert \cB_\cA \vert $ in most cases.
However, with less auxiliary data, the ratio has to be treated more carefully.
For example, smaller $ \vert \cD_\cA \vert $ but larger $ \vert \cB_\cA \vert $ may cause overfitting auxiliary samples and hurt the overall result.

\noindent\textbf{Resampling strategies.}
The target dataset is usually balanced across classes.
The nice property doesn't necessarily hold for the auxiliary dataset.
Thus we propose several resampling strategies.
From Table~\ref{tab:resample}, we see that simple techniques to tailor the distribution into a more balanced one yield nontrivial improvements.

\section{Conclusion}

In this work, we propose OmniSource, a simple yet effective framework for webly-supervised video recognition.
Our method can utilize web data from multiple sources and formats by transforming them into a same format. 
In addition, our task-specific data collection is more data-efficient. 
The framework is applicable to various video tasks.
Under all settings of pretraining strategies, we obtain state-of-the-art performance on multiple benchmarks.

\noindent\textbf{Acknowledgment} This work is partially supported by the SenseTime Collaborative Grant on Large-scale Multi-modality Analysis (CUHK Agreement No. TS1610626 \& No. TS1712093), the General Research Fund (GRF) of Hong Kong (No. 14203518 \& No. 14205719), and Innovation and Technology Support Program (ITSP) Tier 2, ITS/431/18F.


\section*{Appendix: Datasets}

In the main paper, we conduct experiments on three benchmarks, namely Kinetics-400, Youtube-car and UCF101. 
The detailed statistics of the target and auxiliary datasets are listed in Table \ref{tab:dataset_stat}.
Our framework is very data efficient, comparing to approaches which use billion of images, dozens of millions of videos for pretraining. 
All the web data we collected are only several Tera-Bytes. 
After filtering, remaining web data only takes around 3TB in space, which can easily fit into one hard drive. 
In stark comparison, the space required by~\cite{ghadiyaram2019large} is estimated to be at least 100TB.
In this section, we visualize videos in these three datasets, and data in the auxiliary datasets we construct, to show why OmniSource benefits these tasks in different levels.


\begin{table}
	\tiny
	\center
	\caption{\textbf{Dataset Statistics.} Here we show the statistics of dataset we use in our experiments. We report storage amount of lowest cost format for videos (videos when using high fps for training, and frames when using low fps for training). Our framework is data efficient, the amount of data we used is two orders less than web data pretraining approach. Tri-vid denotes trimmed videos and Unt-vid denotes untrimmed videos. }
	\label{tab:dataset_stat}
	\begin{tabular}{c|c|c|c||c|c|c|c|c} 
		\hline \hline
		Target Dataset                & Type                             & Training Size               & Storage                                             & Source Dataset  & Type            & Raw size   & Raw storage                                              & Clean Size                                               \\ \hline \hline
		\multirow{4}{*}{Kinetics-400} & \multirow{4}{*}{Tri-Vid}   & \multirow{4}{*}{\begin{tabular}[c]{@{}c@{}}240K\\ 40K mins \end{tabular}} &  \multirow{4}{*}{140 GB} & GG-k400 & Img           & 6M    & 350 GB   & 2M        \\ \cline{5-9}
		&                                  &     &                                                                     & IG-img & Img           & 7.4M       & 450 GB                                              & 1.5M                                                     \\ \cline{5-9}
		&      &                            &                                                                          & IG-vid & Tri-Vid   & \begin{tabular}[c]{@{}c@{}}1.1M\\ 480K mins\end{tabular} & 1.74 TB & \begin{tabular}[c]{@{}c@{}}500K\\ 250K mins\end{tabular} \\ \cline{5-9}
		&       &                           &                                                                          & k400-untrim    & Unt-Vid & 670K mins                  & 2.44 TB                               & 500K mins                                                \\ \hline \hline
		\multirow{2}{*}{Youtube-car}  & \multirow{2}{*}{Unt-Vid} & \multirow{2}{*}{\begin{tabular}[c]{@{}c@{}}10K\\ 21K mins\end{tabular}} & \multirow{2}{*}{92 GB} & GG-car & Img           & 70K                 &12 GB                                     & 50K                                                      \\ \cline{5-9}
		&                 &                 &                                                                          & YT-car-17k & Unt-Vid & \begin{tabular}[c]{@{}c@{}}28K\\ 63K mins\end{tabular}  & 66 GB & \begin{tabular}[c]{@{}c@{}}17K\\ 38K mins\end{tabular}   \\ \hline \hline
		UCF101                        & Tri-Vid                    & \begin{tabular}[c]{@{}c@{}}10K\\ 1.2K mins\end{tabular}    & 7 GB              & GG-UCF   & Img           & 200K                      & 12 GB                               & 100K                                \\              \hline \hline     
	\end{tabular}
\end{table}

\begin{wrapfigure}{r}{0.45\linewidth}
	\begin{center}
		\includegraphics[width=1\linewidth]{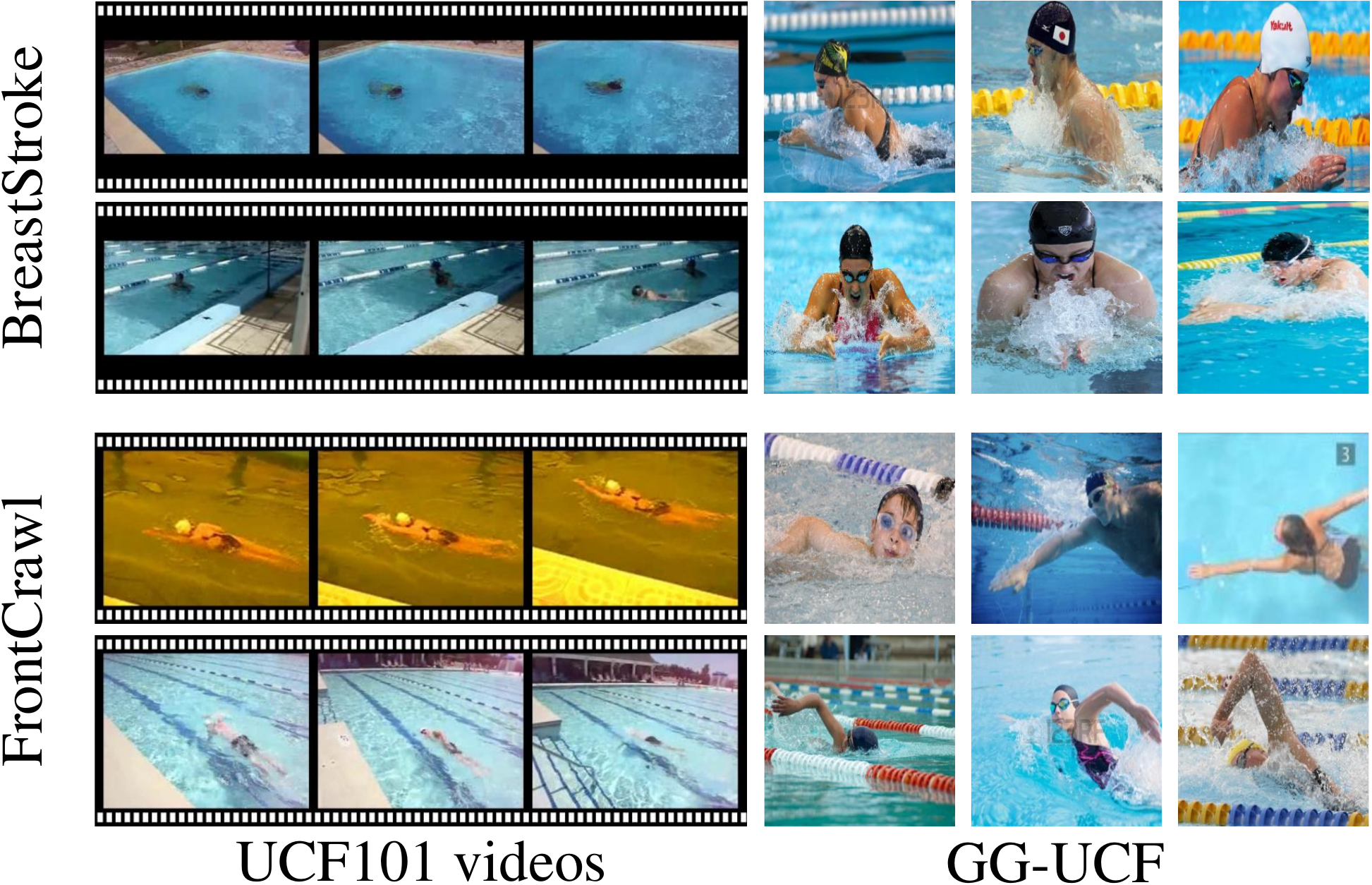}
	
	\caption{\textbf{UCF101.} Visualization of data in UCF101 and its auxiliary dataset. For some classes, web data are more diversified and contain more discriminative poses.}
	\label{fig:ucf}
\end{center}
\end{wrapfigure}

\textbf{Youtube-Car} 
Youtube-Car is the benchmark on which our framework benefits most. 
It mainly has two reasons:
(1) The web data are much cleaner: when searching with the name of a car, it is easy to get a bunch of images with little noise, since nothing is ambiguous.
(2) The source for both target and auxiliary dataset is YouTube, which mean the domain gap is much smaller.
Some samples from Youtube-Car and its auxiliary datasets are visualized in Fig.~\ref{fig:car}.

\textbf{UCF101} Our framework also works on UCF101, which is a small-scale video recognition dataset. UCF101 has much less data diversity and lower visual quality, while auxiliary web data can be complementary in these two aspects. For example, from Fig.~\ref{fig:ucf}, one can hardly tell the difference between BreastStroke and FrontCrawl videos in UCF101. The difference is much more significant in web data. Using our framework, models can learn those discriminative features from web data, and can better recognize videos in the target dataset.

\begin{figure}
	\begin{center}
		\includegraphics[width=1\linewidth]{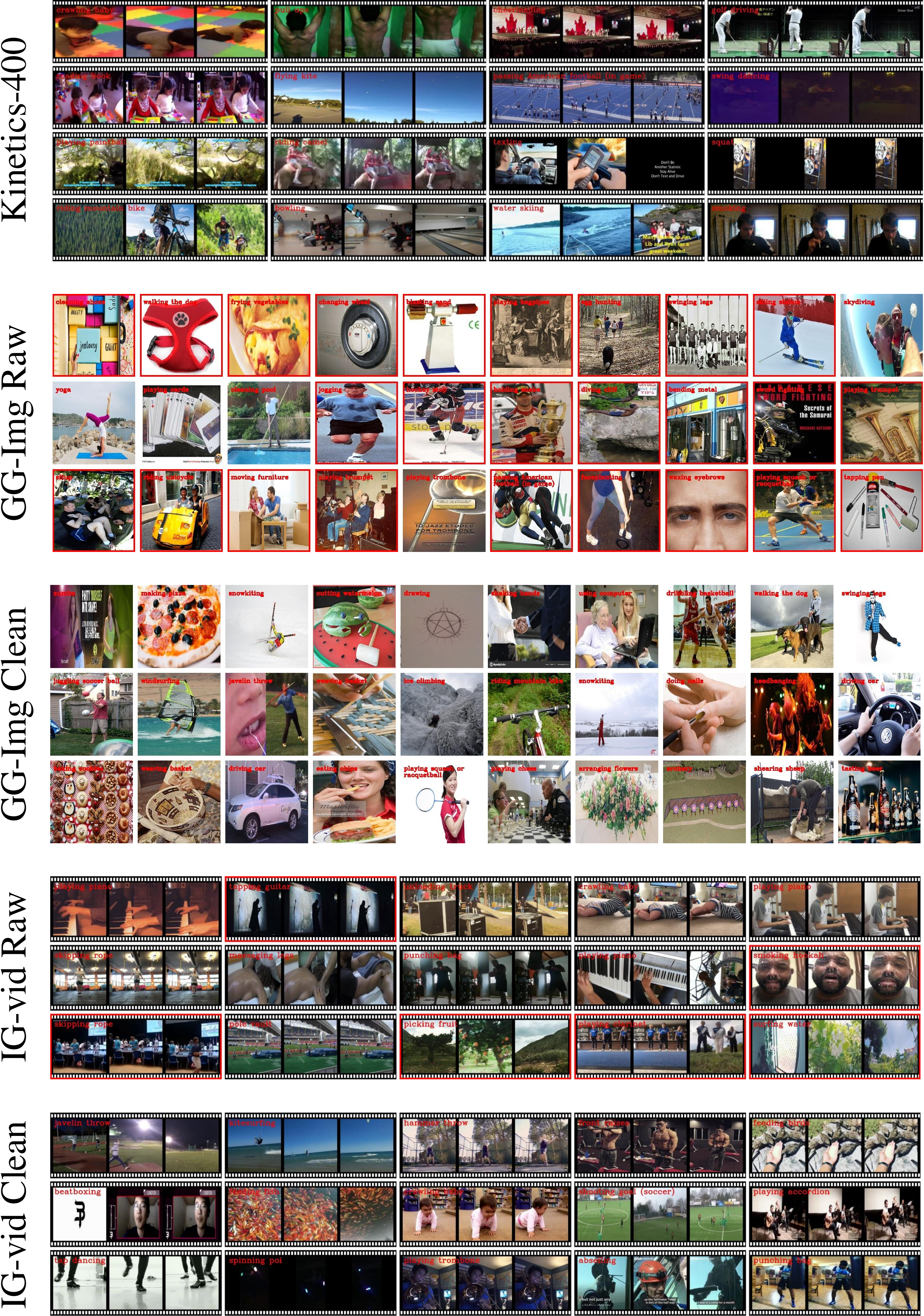}
	\end{center}
	\caption{\textbf{Kinetics.} Data from Kinetics and data from auxiliary datasets are visualized, both raw and clean. Red boxes denote that the image is identified as negative by teacher. There might be some false-negative during teacher filtering, but the data filtered out by teacher are almost clean.}
	\label{fig:kinetics}
\end{figure}

\textbf{Kinetics-400} 
We visualize some images in \texttt{GG-k400} and some videos in \texttt{k400-tr}, \texttt{IG-vid} in Fig.~\ref{fig:kinetics}. 
The observations are summarized below: 
(1) Web data have much more diverse appearance comparing to the target dataset. 
(2) Web data are very noisy. 
The teacher network filtering results reveal that around 60\% - 70\% data in the web data is irrelevant to the task we are interested in. 
(3) We can eliminate noise in web data at the minimal cost of dropping some false negative samples, resulting in a much cleaner auxiliary dataset.

\noindent
\begin{minipage}{\linewidth}
	\begin{minipage}{0.48\linewidth}
		\begin{center}
			\includegraphics[width=\linewidth]{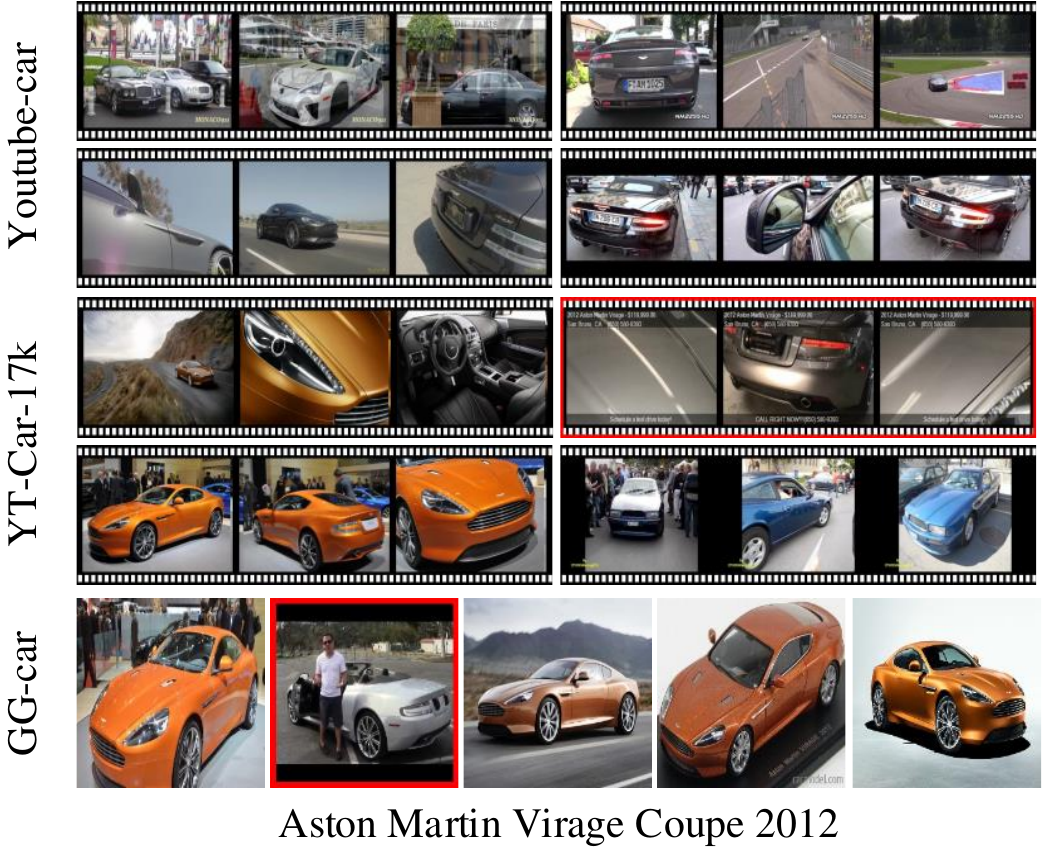}
		\end{center}
		
	\end{minipage}
	\hfill
	\begin{minipage}{0.48\linewidth}
		\begin{center}
			\includegraphics[width=\linewidth]{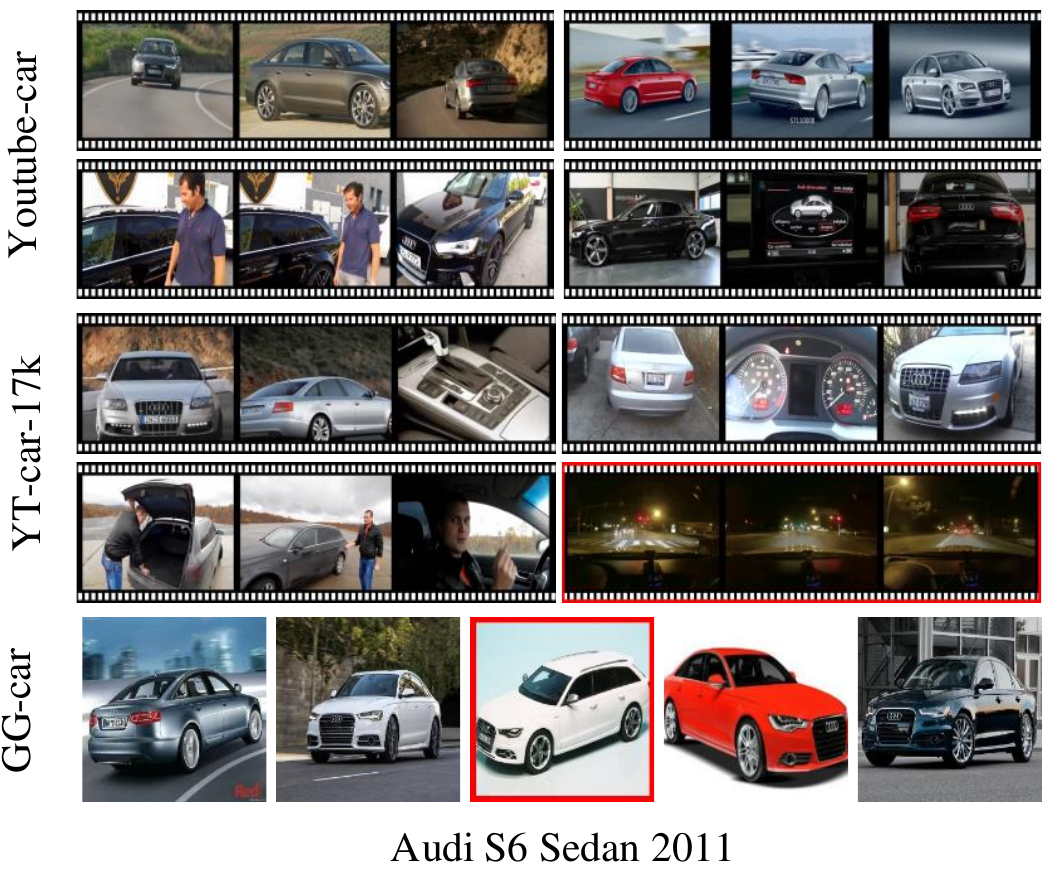}
		\end{center}
	\end{minipage}
	\captionof{figure}{\textbf{Youtube-car.} Visualization of data in Youtube-car and its auxiliary dataset. Since one can easily get images of centain types of cars by querying its name, the quality of the auxiliary dataset is much better. The high quality web data leads to considerable gain in model performance.}
	\label{fig:car}
\end{minipage}

\section*{Appendix: Implementation Details}
Here, we report the implementation details for all our
experiments for Kinetics-400 and transfer learning in UCF101 and HMDB51.

\subsection*{Experiments on Kinetics-400}
For all experiments on Kinetics-400, we use an SGD with momentum of 0.9, and weight decay of $10^{-4}$.
The initial learning rate (LR) we use linearly scales with the number of samples and is decreased to its $ 10^{-1} $.
For TSN-2D experiments, we use $4 \times 10^{-5}$/sample as the starting LR.
The training process lasts 100 epoches and LR decays at 40 and 80 epochs.
For 3D-ConvNet experiments, we use $1.6\times 10^{-4}$/sample as the starting LR for experiments with ImageNet-pretrain, $1.6\times 10^{-3}$/sample as the starting LR for train-from-scratch experiments.
For ImageNet-pretrain experiments, training lasts 150 epochs and LR decays at 90 and 130 epochs.
For train-from-scratch experiments,
we use CosineLR schedule instead of StepLR schedule, and training lasts for 256 epoches and 196 epoches respectively for SlowOnly-4x16 and SlowOnly-8x8, same as training schedules used in \cite{feichtenhofer2019slowfast}.
For IG-65M pretrained irCSN-152, we use $5\times 10^{-6}$/sample as the starting LR.
The training process lasts 58 epochs and LR decays at 32 and 48 epochs, which is consistent with \cite{ghadiyaram2019large}.
Warmup is also used in our experiments, which lasts 34 epochs for the train-from-scratch SlowOnly approach, 16 epochs for irCSN-152.
During warmup, learning rate grows linearly from 0 to the starting LR. The warmup schedules follows~\cite{feichtenhofer2019slowfast,ghadiyaram2019large}.

\subsection*{Experiments for Transfer Learning on UCF-101 and HMDB-51}
We use one simple schedule for all transfer learning experiments.
We we use an SGD with momentum of 0.9, and weight decay of $10^{-4}$.
The starting LR is set to $5\times 10^{-6}$/sample.
We train 90 epoches on UCF101 and HMDB51 and the first 20 epoches are used for warmup, during which learning rate grows linearly from 0 to the starting LR.
No LR decay is performed during training.

\section*{Appendix: More Detailed Experimental Resutls}

Due to space limitation, some experiment results are not described in detail in the main paper. In this part, we discuss these experiments at length.

\subsection*{Verifying the efficacy of OmniSource. }
\noindent\textbf{Why do we need teacher filtering and are search results good enough?}
In the main text, we argue that directly using collected web data for joint training leads to a significant performance drop (Top-1 Accuracy: 70.6\% to 67.4\%) on TSN, which proves the necessity of having a teacher network.
However, since we crawl Top 1000 images for each class name from search engines, one may argue that too many queries lead to bad data quality.
In response to this question, we construct two subset of \texttt{GG-k400-Raw}, which include Top \sfrac{1}{4} (\texttt{GG-k400-Raw-\sfrac{1}{4}}) and Top \sfrac{1}{2} (\texttt{GG-k400-Raw-\sfrac{1}{2}}) results in \texttt{GG-k400-Raw} respectively.
To make sure web images are much more than trimmed videos in the target dataset, we construct a subset of \texttt{k400-tr}, named \texttt{k400-tr-half}, which includes half classes and half videos per class.
We jointly train \texttt{k400-tr-half} with different auxiliary datasets.
From Table~\ref{tab:teacher_filtering}, we see that raw web data are of low quality, even for top search results. 
Thus teacher filtering is an essential step in OmniSource.

\begin{table}
	\caption{Joint training \texttt{k400-tr-half} with different raw web datasets. We see that even top search results are of bad quality, lead to inferior performance. Thus teacher filtering is essential }
	\label{tab:teacher_filtering}
	\centering
	\begin{tabular}{c|c|c|c}
		\hline \hline
		Target Dataset                & Source Dataset  & Top-1 & Top-5 \\ \hline \hline
		\multirow{4}{*}{\texttt{k400-tr-half}} & \texttt{/}              & 72.2  & 90.3  \\ \cline{2-4}
		& \texttt{GG-k400-Raw}     & 70.3  & 89.2  \\ \cline{2-4}
		& \texttt{GG-k400-Raw-\sfrac{1}{2}} & 69.8  & 88.7  \\ \cline{2-4}
		& \texttt{GG-k400-Raw-\sfrac{1}{4}} & 69.9  & 88.7  \\ \hline \hline
	\end{tabular}
\end{table}

\noindent\textbf{Does every data source contribute? }
In the main text, we use two groups of experiments which use ImageNet pretrained TSN-3seg-R50 and SlowOnly-4x16-R50 as baselines, to prove that every source contributes.
Besides that, the conclusion also holds for SlowOnly-4x16-R50 trained from scratch.
From Table~\ref{tab:omnisource}, we see that for the train-from-scratch setting, each data source not only contributes to the target task, but the improvement is much larger than the ImageNet-pretrain setting.

\begin{table}
	\scriptsize
	\begin{center}
		\caption{For the train-from-scratch setting, every data source also contributes to the target task. The improvement is much larger compared to the ImageNet-pretrain setting. (FT: ImageNet-pretrain; SC: train-from-scratch)}
		\label{tab:omnisource}
		\begin{tabular}{c||c||c|c|c|c||c}
			\hline \hline
			
			Arch/Dataset  & \texttt{K400-tr}  & +\texttt{GG-k400} & +\texttt{GG\&IG-img} & +\texttt{IG-vid} & +\texttt{K400-untr} & + All \\
			\hline \hline
			\begin{tabular}[c]{@{}c@{}}SlowOnly \\4x16, R50 {[}FT{]} \end{tabular} & 73.8/90.9 & 74.5/91.4 & 75.2/91.6 & 75.2/91.7 & 74.5/91.1  & 76.6/92.5 \\ \hline
			\begin{tabular}[c]{@{}c@{}}SlowOnly\\4x16, R50 {[}SC{]}\end{tabular} & 72.9/90.9 & 74.1/91.0 & 74.8/91.4 & 75.8/92.0 & 74.8/91.2  & 76.8/92.5 \\
			\hline \hline
		\end{tabular}
	\end{center}
\end{table}

\noindent\textbf{Do features learned by OmniSource transfer to other tasks?} In this section, we provide extensive experiment results on transfer learning, much more than results presented in the main text.
Table~\ref{tab:transfer} lists transfer learning results on UCF101-split1 and HMDB-split1.
Those results further support 2 points proposed in the main text: (1) OmniSource framework can learn better representation, which leads to significant performance improvement on downstream tasks.
(2) ImageNet-pretraining is not indispensable for OmniSource to learn good representation.
When combined with flow stream, state-of-the-art results on UCF101 and HMDB51 can be achieved by finetuning models jointly trained on Kinetics and auxiliary datasets.
Table~\ref{tab:transfer_sota} compares the transfer learning performance of OmniSource trained models with other state-of-the-art approaches.
We see that OmniSource outperforms other methods by a large margin.

\begin{wraptable}{r}{.5\linewidth}
	\caption{We explore different combinations to build a 3-frame snippet, and find that 1 Pos. + 2 Neg. is the best choice.}
	\label{tab:untr2snip}
	\begin{center}
		\begin{tabular}{l|l|l}
			\hline \hline
			Configuration   & Top-1 & Top-5 \\ \hline \hline
			3 Rand.         & 71.42 & 89.34 \\ \hline
			3 Pos.          & 71.22 & 89.54 \\ \hline
			2 Pos. + 1 Neg. & 71.44 & 89.57 \\ \hline
			1 Pos. + 2 Neg. & 71.66 & 89.63 \\ \hline \hline
		\end{tabular}
	\end{center}
\end{wraptable}

\noindent\textbf{Untrimmed videos to snippets. }
In the main paper, we mention that combining negative frames and positive frames is a good practice to construct harder snippets,  which leads to better recognition performance. 
We provide detailed results in Table~\ref{tab:untr2snip}, in which we explore each possible combinations during joint training \texttt{k400-tr} and \texttt{k400-untr} with TSN-3seg-R50 baseline. 
We find that combining one positive frame and two negative frames to form a 3-frame snippet leads to best performance.

\begin{table}
	\centering
	\caption{Detailed results of transfer learning. We report Top-1 accuracies on the official split-1 of UCF101 and HMDB51. We see that OmniSource framework can learn better representation which transfers to other recognition tasks well, even without ImageNet pretraining. }
	\label{tab:transfer}
	\vspace{2mm}
	\begin{tabular}{c|c|c|c|c}
		\hline \hline
		Architecture                                                                      & w/. ImageNet-pretrain & w/. OmniSource & UCF101-Top1 & HMDB51-Top1 \\ \hline \hline
		\multirow{2}{*}{\begin{tabular}[c]{@{}c@{}}TSN-3seg\\ ResNet50\end{tabular}}      & $\checkmark$          &                & 91.51       & 63.53       \\ \cline{2-5} 
		& $\checkmark$          & $\checkmark$   & 93.29       & 65.88       \\ \hline \hline
		\multirow{2}{*}{\begin{tabular}[c]{@{}c@{}}TSN-3seg\\ Efficient-b4\end{tabular}}  & $\checkmark$          &                & 92.52       & 66.27       \\ \cline{2-5} 
		& $\checkmark$          & $\checkmark$   & 93.05       & 66.54       \\ \hline \hline
		\multirow{4}{*}{\begin{tabular}[c]{@{}c@{}}SlowOnly-4x16\\ ResNet50\end{tabular}} & $\checkmark$          &                & 94.69       & 69.35       \\ \cline{2-5} 
		& $\checkmark$          & $\checkmark$   & 95.98       & 70.71       \\ \cline{2-5} 
		&                       &                & 94.05       & 65.82       \\ \cline{2-5} 
		&                       & $\checkmark$   & 96.01       & 70.98       \\ \hline \hline
		\multirow{4}{*}{\begin{tabular}[c]{@{}c@{}}SlowOnly-8x8\\ ResNet101\end{tabular}} & $\checkmark$          &                & 96.40       & 76.41       \\ \cline{2-5} 
		& $\checkmark$          & $\checkmark$   & 97.38       & 78.95       \\ \cline{2-5} 
		&                       &                & 96.61       & 75.82       \\ \cline{2-5} 
		&                       & $\checkmark$   & 97.52       & 79.02       \\ \hline \hline
		
	\end{tabular}
\end{table}

\subsection*{Validating the good practices in OmniSource}
\noindent\textbf{Impact of teacher choice. } 
In the main paper, we mention that for web video data, 3D teachers always outperform 2D ones. 
Besides that, the conclusion that the accuracy of the student network increases when a better teacher network is used also holds for web video data.
Here, we provide some quantitative results to prove those conclusions in Table~\ref{tab:better_teacher}. 
SlowOnly-4x16-R50 with ImageNet-pretrain is used as the student network.

\begin{table}
	\normalsize
	\centering
	\caption{We compare transfer learning results with state-of-the-art approaches. We report mean Top-1 accuracies on three splits of UCF101 and HMDB51. We see that OmniSource framework not only outperforms RGB-Only methods. When fused with the flow stream, it surpasses all methods by a large margin, even for those which ensemble results of RGB, Flow and other modalities (*We reimplement Flow-I3D as our flow stream) }
	\label{tab:transfer_sota}
	\vspace{2mm}
	\begin{tabular}{c|c|c|c}
		\hline \hline
		Model                    & Pretrain              & UCF101 & HMDB51 \\ \hline \hline
		Two-Stream    \cite{simonyan2014two}           & ImageNet              & 88.0   & 59.4   \\ \hline
		TSN    \cite{wang2018temporal}                  & ImageNet              & 94.2   & 69.4   \\ \hline \hline
		RGB-I3D\cite{carreira2017quo}                  & ImageNet + Kinetics   & 95.6   & 74.8   \\ \hline
		Flow-I3D\cite{carreira2017quo}                 & ImageNet + Kinetics   & 96.7   & 77.1   \\ \hline
		Two-Stream-I3D\cite{carreira2017quo}           & ImageNet + Kinetics   & 98.0   & 80.7   \\ \hline \hline
		I3D + PoTion\cite{choutas2018potion}             & ImageNet + Kinetics   & 98.2   & 80.9   \\ \hline
		I3D + PA3D\cite{yan2019pa3d}               & ImageNet + Kinetics   &  /     & 82.1   \\ \hline \hline
		SlowOnly-8x8-R101        & Kinetics + OmniSource & \textbf{97.3}   & \textbf{79.0}   \\ \hline
		SlowOnly-8x8-R101 + Flow$^1$ & Kinetics + OmniSource & \textbf{98.6}   & \textbf{83.8}   \\ \hline \hline
	\end{tabular}
\end{table}

\begin{table}
	\normalsize
	\centering
	\caption{More results on the impact of teacher choice. 3D teachers always outperform 2D ones. The accuracy of the student network increases when a better teacher network is used. }
	\label{tab:better_teacher}
	\vspace{2mm}
	\begin{tabular}{c|c|c|c|c|c}
		\hline \hline
		Aux. Dataset               & Teacher           & Teacher Top-1 & 2D / 3D ? & Top-1 & Top-5 \\ \hline \hline
		\multirow{3}{*}{IG-vid}    & TSN-3seg-R50      & 70.6          & 2D        & 73.2  & 90.8  \\ \cline{2-6} 
		& SlowOnly-4x16-R50 & 73.8          & 3D        & 75.2  & 91.7  \\ \cline{2-6} 
		& IRCSN-152         & 82.6          & 3D        & 75.4  & 91.9  \\ \hline \hline
		\multirow{3}{*}{K400-untr} & TSN-3seg-R50      & 70.6          & 2D        & 74.1  & 91.0  \\ \cline{2-6} 
		& SlowOnly-4x16-R50 & 73.8          & 3D        & 74.5  & 91.1  \\ \cline{2-6} 
		& IRCSN-152         & 82.6          & 3D        & 75.0  & 91.4  \\ \hline \hline
	\end{tabular}
\end{table}


\section*{Appendix: Improvement Analysis}
We further study the improvement of our framework, when using the full auxiliary set for training. Recall that our framework can improve 3.0\% and 3.9\% respectively on 2D and 3D baseline with all auxiliary data we collected, We analyze the improvement on confusing pairs over these two cases. We use delta of confusion score ($\Delta_{ij}$) to denote the improvement:

\begin{equation}
\Delta_{ij} = Oscore_{ij} - Bscore_{ij},
\end{equation}

where $Oscore_{ij}$ denotes the confusion score of pair $<i, j>$ when trained with OmniSource, and $Bscore_{ij}$ denotes the confusion score of pair $<i, j>$ of baseline model.

We show success and failure cases of 2D model in Table~\ref{tab:conf-2d}.
The contribution of our framework mainly attributes to the better object recognition ability.
Besides that, it also improves when discriminative element can be found in web data, like two hands touched in handshaking, two head touched in headbutting, etc..
There are also failure cases when motion is needed for action recognition or when the taxonomy is not reasonable.

We show success and failure cases of 3D model in Table~\ref{tab:conf-3d}.
Thanks to the capability of using motion cues for action recognition, the pair 'rock scissors paper' and 'slapping' is no longer a failure case ($ \Delta $ from +0.176 to -0.059).
However, when appearance and motion are all similar, our framework might fail due to the introduced noises.

\begin{figure}
	\begin{center}
		\includegraphics[width=1\linewidth]{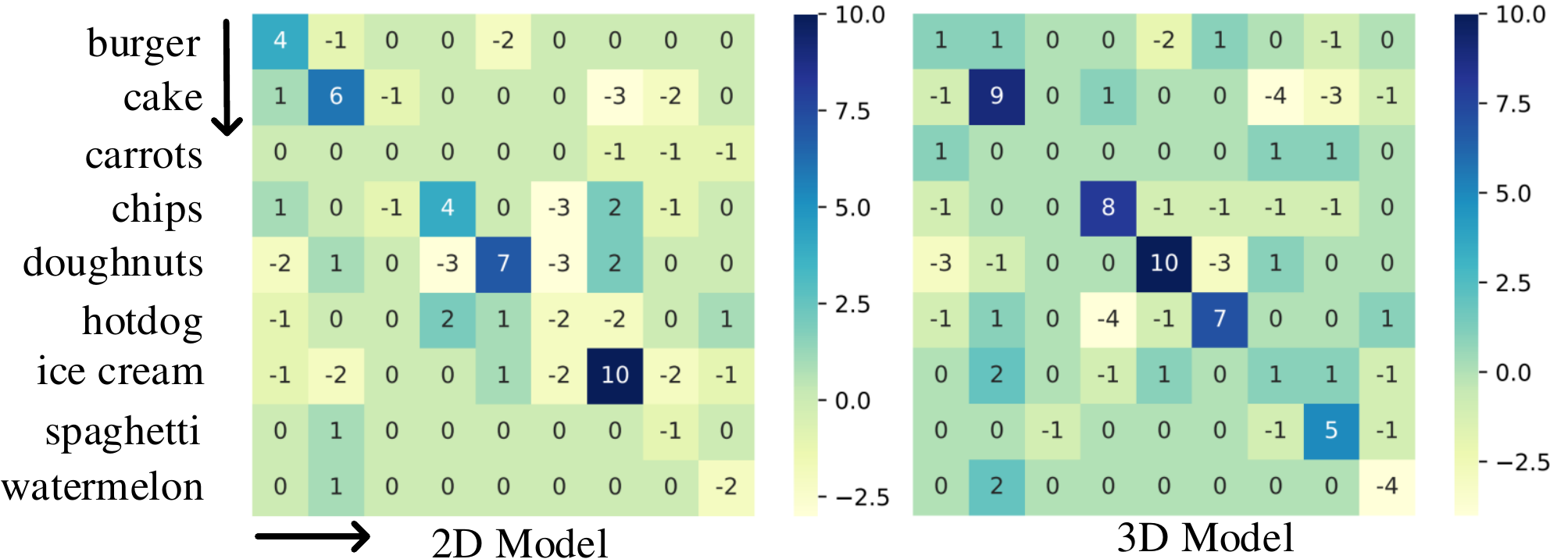}
	\end{center}
	\caption{\textbf{Improvement on eating something.} Rows denote groundtruth and columns denote predictions. Block$_{ij}$ represents the difference in numbers of samples which belongs to class $i$ but recognized as class $j$ between the baseline and our model.}
	\label{fig:confuse}
\end{figure}

Due to the improved ability of object recognition, the accuracy improvement on actions of eating something is much more significant.
On average, the accuracy for eating something improved 5.8\%, 8.3\% for 2D and 3D models respectively, while the average improvement for all classes are 3.0\% and 3.9\%.
We visualize the improvement on this subset in Fig.~\ref{fig:confuse}.

\noindent
\begin{minipage}{\linewidth}
	\begin{minipage}{.48\linewidth}
	\begin{center}
		\tiny
		\begin{tabular}{c|lll}
			\hline
			\multicolumn{1}{l}{Case} & Action 1            & Action 2         & $\Delta_{ij}$ $\downarrow$ \\ \hline
			\multirow{5}{*}{Success} & rock scissors paper & shaking hands    & -0.160 \\ \cline{2-4} 
			& headbutting         & sniffing         & -0.159 \\ \cline{2-4} 
			& sweeping floor      & mopping floor    & -0.113 \\ \cline{2-4} 
			& eating chips        & eating doughnuts & -0.103 \\ \cline{2-4} 
			& eating ice creams   & eating cake      & -0.100 \\ \hline
			\multirow{2}{*}{Failure} & rock scissors paper & slapping         & +0.176 \\ \cline{2-4} 
			& drinking              & drinking shots       & +0.158 \\ \hline
		\end{tabular}
	\end{center}
	\captionof{table}{Confusion Score Delta for 2D models. Lower delta means larger gain in discriminative power of these two classes. Top-5 and Lowest-2 entries are displayed. }
	\label{tab:conf-2d}
	\end{minipage}
	\hfill
	\begin{minipage}{.48\linewidth}
		\begin{center}
			\tiny
			\begin{tabular}{c|lll}
				\hline
				\multicolumn{1}{l}{Case} & Action 1         & Action 2         & $\Delta_{ij}$ $\downarrow$ \\ \hline
				\multirow{5}{*}{Success} & slapping         & headbutting      & -0.235 \\ \cline{2-4} 
				& eating doughnuts & eating hotdog    & -0.153 \\ \cline{2-4} 
				& eating chips     & eating hotdog    & -0.121 \\ \cline{2-4} 
				& faceplanting     & drop kicking     & -0.120 \\ \cline{2-4} 
				& cooking chicken  & cooking sausages & -0.110 \\ \hline
				\multirow{2}{*}{Failure} & baking cookies   & making a cake    & +0.119 \\ \cline{2-4} 
				& yawning          & sneezing         & +0.104 \\ \hline
			\end{tabular}
		\end{center}
		\captionof{table}{Confusion Score Delta for 3D models. Lower delta means larger gain in discriminative power of these two classes. Top-5 and Lowest-2 entries are displayed. }
		\label{tab:conf-3d}
	\end{minipage}	
\end{minipage}

{\small
\bibliographystyle{ieee_fullname}
\bibliography{egbib}
}

\end{document}